\documentclass[times, review, 10pt]{elsarticle}

\usepackage{amsmath,amssymb}
\usepackage{graphicx}
\usepackage{booktabs,tabularx}
\usepackage{setspace}
\usepackage{caption}
\usepackage{geometry}
\usepackage{hyperref}
\usepackage{etoolbox}

\makeatletter
\patchcmd{\pprintMaketitle}
  {\Large\@title}
  {\fontsize{18pt}{18pt}\selectfont\@title}
  {}{\PackageError{ExpertHTR}{Could not set the title to 14 pt}{}}
\patchcmd{\pprintMaketitle}
  {\normalsize\elsauthors}
  {\normalsize\elsauthors}
  {}{\PackageError{ExpertHTR}{Could not set author names to 8 pt}{}}
\patchcmd{\MaketitleBox}
  {\Large\@title}
  {\fontsize{14pt}{14pt}\selectfont\@title}
  {}{\PackageError{ExpertHTR}{Could not set the title to 14 pt}{}}
\patchcmd{\MaketitleBox}
  {\normalsize\elsauthors}
  {\normalsize\elsauthors}
  {}{\PackageError{ExpertHTR}{Could not set author names to 8 pt}{}}
  \def\ps@pprintTitle{%
  \let\@oddhead\@empty
  \let\@evenhead\@empty
  \let\@oddfoot\@empty
  \let\@evenfoot\@oddfoot
}
\makeatother

\biboptions{numbers,sort&compress}

\newcolumntype{Y}{>{\raggedright\arraybackslash}X}
\begin{document}
\begin{frontmatter}

\title{ExpertHTR: Structure-Aware Multitask Supervision and Pointer-Guided Fine-Grained Experts for Page-Level Handwritten Text Recognition}

\author[1,2]{Dang Hoai Nam}
%\ead{23520967@gm.uit.edu.vn}
\author[1,2]{Nguyen Duy Hieu}
%\ead{24520501@gm.uit.edu.vn}
\author[3]{Quang Huu Hieu}
%\ead{hieuquang@aj-tech.jp}
\author[1,2,4]{Vo Nguyen Le Duy\corref{cor1}}
\ead{duyvnl@uit.edu.vn}

\address[1]{University of Information Technology, Ho Chi Minh City, Vietnam}
\address[2]{Vietnam National University, Ho Chi Minh City, Vietnam}
\address[3]{AJ Technologies, Nagoya, Japan}
\address[4]{RIKEN Center for Advanced Intelligence Project, Tokyo, Japan}
\cortext[cor1]{Corresponding author.}

\begin{abstract}
Vision--language models (VLMs) offer a single generative interface for
page-level handwritten text recognition (HTR), but two obstacles limit them.
Annotated handwriting collections are small and heterogeneous, and the visual
tokens of a VLM are coarse: after patch merging, each token typically
summarizes a region of about $32\times32$ pixels. We propose ExpertHTR, which
addresses both. At the supervision level, annotations from seven collections
are mapped to a common Page--Region--Line representation, from which four
complementary tasks are derived without additional labels: Page
Transcription, Physical Line Counting, Text-to-Line Localization, and
Line-Span Transcription. At the model level, a pointer-guided fine-grained
Mixture-of-Experts (PF-MoE) enables selected decoder layers to retrieve
fine-grained visual features: a pointer initialized from the layer's own
attention selects, for every generated token, multi-scale convolutional
cells at 8-, 16-, and 32-pixel strides, and routed experts inject these
features through a gated correction without adding visual tokens. On 1,327
test pages from seven modern and historical benchmarks in six languages, a
single Qwen3.5-0.8B checkpoint reaches a character error rate (CER) of
6.81\%, compared with 8.81\% for the same multitask model without PF-MoE,
and 2.96\% CER on IAM, compared with 4.16\% for the best published
specialized system. With a churro-3B backbone, ExpertHTR obtains the lowest
Bentham CER of all evaluated systems (2.79\%). Ablations and analyses show
that the fine-grained features mainly stabilize long generation, reducing
skipped lines, repeated lines, and repetition loops, and help identify
cancelled and illegible text. Without supervision, the pointer concentrates
on the finest stride. The code is available at \url{https://github.com/DAIR-Group/ExpertHTR}.
\end{abstract}

% Research highlights
% \begin{highlights}
% \item One vision-language recognizer reads full pages from seven handwriting datasets.
% \item Four auxiliary tasks are derived from existing page annotations without new labels.
% \item Decoder attention guides a pointer to fine 8--32\,px image cells for each output token.
% \item Pointer-guided experts cut CER from 8.81\% to 6.81\%; looping pages drop from 16 to 3.
% \item IAM CER is 2.96\% vs. 4.16\% for specialized systems; best Bentham CER is 2.79\%.
% \end{highlights}

% Keywords
% Each keyword is seperated by \sep

\end{frontmatter}

\doublespacing

\section{Introduction}
\label{sec:introduction}

Handwritten text recognition (HTR) has moved from isolated words and lines to
paragraphs and complete pages \citep{span_htr,dan_htr,fasterdan_htr,metadan_htr}.
Page-level systems learn character recognition, line transitions, and reading
order in a single model, without a separate line-segmentation stage. This
setting also brings requirements that matter little for single lines: the
recognizer must cover every line of the page, keep the physical line breaks,
follow the reading order across separate text blocks, and remain reliable over
outputs of several hundred tokens. These requirements are hardest to meet on
historical and degraded documents and on collections that differ from the
training data \citep{htr_generalization}. At the same time, annotated
page-level collections are small and differ in language, script, period,
layout, and annotation conventions, which is why document-level recognizers
usually depend on line-level pretraining or staged training before they are
trained on complete pages \citep{span_htr,dan_htr}.

Pretrained vision--language models (VLMs) are an attractive starting point:
any recognition task can be posed as image-to-text generation, and
large-scale pretraining supplies broad visual and linguistic knowledge.
General multimodal models are, however, unreliable on handwriting, especially
across languages and historical periods \citep{crosilla_htr}, and become
strong recognizers only after targeted adaptation \citep{churro_htr}. When we
fine-tune a compact VLM for page-level HTR, two kinds of errors remain. First,
the model loses track of the page structure: it omits lines, transcribes a
line twice, or falls into a repetition loop that runs until the token limit. With the page transcription as the only target, such failures are
penalized only indirectly, through a token-level loss over a long sequence.
Second, it often writes words that fit the context but not the strokes, such
as a common word instead of a rare one, or a dictionary word instead of a
cancelled one. We hypothesized that this second kind of error is partly
caused by the coarse visual input. To keep pages tractable, the VLM merges
neighboring patches, so that each visual token summarizes a
$32\times32$-pixel region, whereas diacritics, cancellation strokes, and the
components of Chinese characters often occupy only a few pixels of it.
Raising the input resolution would recover this detail, but it would multiply
the number of visual tokens for the whole page, although the detail is needed
only where the model is currently reading.

We propose ExpertHTR, which addresses these two problems at two levels. At
the supervision level, the annotations of seven collections are mapped to a
common Page--Region--Line representation, from which four tasks are derived
without additional labels. Page Transcription remains the main task, while
Physical Line Counting, Text-to-Line Localization, and Line-Span
Transcription turn line coverage, the position of text on the page, and the
boundaries of line ranges into short, explicit targets. At the model level, a
pointer-guided fine-grained Mixture-of-Experts (PF-MoE) lets selected
decoder layers look at the page again, at a finer resolution, where they are
reading. A frozen ConvNeXt~\citep{convnext} with a small local refinement
provides cells at 8-, 16-, and 32-pixel strides. For every generated token, a
pointer initialized from the layer's own attention selects a few of these
cells, and routed experts add the retrieved evidence to the hidden state
through a learned gate. We use routed experts rather than a single additional
layer because the training data are heterogeneous in script, period, and
writing style, and the content ranges from ordinary words to digits and
cancelled passages; routing lets such tokens use different parameters without
increasing the per-token cost. The router never sees language, script,
dataset, or task identities, and the number of visual tokens is unchanged.

On seven benchmarks covering modern and historical handwriting in six
languages and two scripts, a single checkpoint reduces the micro-averaged
character error rate (CER) of the same multitask model without PF-MoE from
8.81\% to 6.81\% and obtains a lower CER on the IAM paragraph-level benchmark
\citep{marti2002iam} than published page-level specialized systems. With a
churro-3B backbone, ExpertHTR obtains the lowest Bentham CER of all evaluated
systems (2.79\%). Used on its own to fine-tune other OCR/VLM models, the
multitask data lowers CER in 34 of 49 model--dataset pairs. The analysis also
qualifies our initial hypothesis: under teacher forcing, removing the
fine-grained evidence leaves the likelihood of ordinary characters unchanged.
The gains come mainly from more stable generation, with fewer skipped lines
and repetition loops, and from better handling of cancelled and illegible
text.

The main contributions of this work are as follows:

\begin{itemize}

    \item A Page--Region--Line representation that unifies the structural
    annotations of heterogeneous HTR collections, and four tasks derived from
    it without additional labeling. At an equal number of updates, they improve
    Page-Transcription-only training in every single-source setting, and
    they also improve fine-tuned OCR/VLM models of other architectures.

    \item PF-MoE, a module that gives a VLM decoder token-level access to
    fine multi-scale features of the page through an attention-initialized
    pointer and routed experts, without adding visual tokens and without
    language, script, dataset, or task identities. We attach it to three
    backbones of different families and sizes, with full fine-tuning and
    with LoRA.

    \item An evaluation on seven benchmarks with component ablations, an
    error decomposition, and an analysis of stride
    selection, expert specialization, and the effect of removing the
    evidence, which shows where the module helps and where it does not. The complete code, including training, evaluation, and inference
    for all three backbones, is publicly available at
    \url{https://github.com/DAIR-Group/ExpertHTR}.

\end{itemize}

\section{Related Work}
\label{sec:related_work}
This section reviews three lines of research most closely related to
ExpertHTR. We first discuss recent progress in paragraph- and page-level HTR,
followed by vision--language models and multitask adaptation for handwriting
recognition. We then review sparse Mixture-of-Experts architectures and
mechanisms that give a model access to fine-grained visual evidence, which
together motivate the PF-MoE design.

\subsection{Page-Level Handwritten Text Recognition}

SPAN introduced
segmentation-free paragraph recognition, while the Document Attention Network
(DAN) extended this direction to complete handwritten documents by learning
transcription, line transitions, and reading order without an explicit
line-segmentation stage \citep{span_htr,dan_htr}. Faster DAN further improved
document-level decoding efficiency through multi-target prediction
\citep{fasterdan_htr}. DANIEL broadened handwritten document recognition
toward document understanding by jointly addressing transcription, layout
analysis, and information extraction across multiple languages and document
types \citep{daniel_htr}. More recently, Meta-DAN proposed a more efficient
prediction strategy for page-level HTR and reported strong results across
several handwritten document benchmarks \citep{metadan_htr}. Our focus differs
from the broader document-understanding setting of DANIEL. Rather than jointly
predicting recognition, layout, and information-extraction outputs, ExpertHTR
derives several supervision tasks from structural information already
available in HTR datasets and uses them to support a common page-level
transcription objective.

Training reliable page-level recognizers remains difficult when annotated
document data are limited. Several representative approaches therefore build
recognition ability in simpler settings before moving to complete documents.
DAN, for example, transfers recognition knowledge learned from synthetic text
lines to document-level training, while MSdocTr-Lite uses a curriculum-based
strategy to progressively move toward full-page recognition
\citep{dan_htr,msdoctr_lite}. A recent HTR survey similarly identifies
line-level supervision, pretraining, curriculum learning, and synthetic data
as recurring strategies for addressing limited annotated handwriting data
\citep{htr_survey}. Related Transformer-based recognition systems have also
explored pretraining to reduce their dependence on task-specific labeled data.
DTrOCR leverages a pretrained generative language model within a decoder-only
recognition architecture, while UCL-MHTR combines self-supervised pretraining
with subsequent multilingual HTR adaptation
\citep{fujitake2024dtrocr,dhiaf2025ucl}.

Data scarcity has also motivated substantial work on handwriting synthesis.
Recent methods include adversarial, diffusion-based, Transformer-based, and
autoregressive approaches for generating diverse handwritten text
\citep{handwriting_synthesis_survey,fwgan,writevit,emuru}. Synthetic handwriting
can expand the visual training distribution when real annotations are scarce,
and recent studies have examined how reliably such data improve HTR under
low-resource conditions \citep{quo_vadis_htg}. Our work addresses the same
data-scarcity problem from a different direction: instead of generating new
handwriting images, we make broader use of the structural information already
available in existing HTR collections.

Generalization across handwriting collections introduces another challenge.
Recognition accuracy can degrade substantially when the training and target
datasets differ in textual or visual characteristics
\citep{htr_generalization}. This issue becomes particularly relevant when one
recognizer is trained across modern and historical handwriting, multiple
languages and scripts, and documents with different structural properties.
ExpertHTR addresses this setting by bringing heterogeneous collections into a
common page-level training framework: their structural annotations are
organized through a Page--Region--Line representation and used to construct
complementary supervision for the same transcription model.

\subsection{Vision--Language Models and Multitask Adaptation}
\label{sec:rw_vlm}

Pretrained vision--language models (VLMs) provide a general image-to-text
generation interface that is well suited to tasks that can be expressed as
conditional text generation. Their broad visual and linguistic representations
also make them an attractive starting point for document recognition.
Nevertheless, general multimodal pretraining does not guarantee reliable
handwriting recognition.

Recent evaluations of multimodal models on handwritten documents report large
differences in recognition quality across languages, historical periods, and
document conditions, with difficult historical handwriting remaining
particularly challenging \citep{crosilla_htr}. Domain-specific adaptation can
substantially improve this behavior. CHURRO (churro-3B), for example, adapts an open-weight
VLM for historical text recognition using a large collection of documents
covering diverse languages, scripts, periods, and document conditions
\citep{churro_htr}. These results suggest that pretrained VLMs provide a useful
starting point, but effective HTR still requires targeted handwriting
supervision.

Multitask adaptation offers another way to provide more informative
supervision than a single final sequence target. Uni-MuMER follows this
direction for handwritten mathematical expression recognition by constructing
complementary tasks that address different aspects of recognition within a
unified VLM fine-tuning framework \citep{unimummer}. Although its task and
output structure differ from page-level HTR, it provides a relevant example
of using complementary generative objectives during VLM adaptation.

Page-level HTR introduces its own structural requirements, including complete
text coverage, physical-line organization, reading order, and reliable
generation over long sequences. ExpertHTR therefore constructs four
HTR-specific objectives---Page Transcription, Physical Line Counting,
Text-to-Line Localization, and Line-Span Transcription---from existing
structural annotations. Unlike broader document-understanding multitask
settings, these objectives are all designed to support the same page-level
transcription problem and require no additional manual labels.

\subsection{Mixture-of-Experts and Fine-Grained Visual Evidence}
\label{sec:rw_moe}

\paragraph{Sparse Mixture-of-Experts.}
Mixture-of-Experts (MoE) layers increase model capacity by combining several
expert transformations with a learned router. Sparse MoE models such as
GShard and Switch Transformer activate only a small subset of experts for each
token, so that total capacity grows without evaluating every expert for every
representation, and use an auxiliary load-balancing loss to keep the experts
in use \citep{gshard,switch_transformer}. Mixtral applies Top-2 routing to the
feed-forward layers of a decoder-only language model \citep{mixtral}, and
DeepSeekMoE adds always-active shared experts that capture common knowledge
while the routed experts specialize \citep{deepseekmoe}. Sparse upcycling
initializes experts from a pretrained dense checkpoint instead of training
them from scratch \citep{sparse_upcycling}. In all of these designs, the
experts receive the same hidden state as the feed-forward layer they replace
or accompany; they add conditional capacity but no new information. PF-MoE
keeps the pretrained feed-forward layer as an always-active branch and adds
Top-1 routed experts initialized to contribute zero, so training starts
exactly from the dense model. Unlike the experts above, each PF-MoE expert
also receives visual evidence retrieved for the token it processes.

\paragraph{Visual token resolution in VLMs.}
VLMs typically reduce the number of visual tokens before the language
decoder. Qwen2-VL, for example, merges each $2\times2$ group of neighboring
patches into one token, which makes high-resolution document images
tractable \citep{qwen2vl}. The resulting tokens are well suited to layout and
large print but summarize regions that are large relative to handwritten
strokes. A common remedy is to increase the input resolution or to encode
additional high-resolution crops \citep{monkey}. This restores detail, but the
number of visual tokens, and with it the attention cost and key--value memory,
grows for the whole page, although fine evidence is needed only where text is
being transcribed at a given step.

\paragraph{Selective access to visual features.}
Several architectures read visual features selectively instead of
uniformly. Gated cross-attention layers in Flamingo let a frozen language
model attend to visual features from inside the decoder
\citep{flamingo}. Deformable attention samples a small set of locations around
a reference point at several feature scales, rather than attending to every
position \citep{deformable_detr}, and feature pyramids provide such multi-scale
features from a convolutional backbone \citep{fpn}. In HTR, attention-based
document recognizers such as DAN attend to the encoder feature map at every
decoding step and thereby read the region that corresponds to the character
being produced \citep{dan_htr}. PF-MoE combines these ideas within a
pretrained VLM. A pointer derived from the decoder's own attention selects a
few fine multi-scale cells around the location being transcribed, a small
cross-attention reads them, and the result is injected through routed
experts. The visual token sequence of the VLM is left unchanged, and no
language, script, dataset, or task identity is used for routing or retrieval.

\section{Proposed Method}
\label{sec:method}

ExpertHTR adapts a pretrained vision--language model (VLM) to page-level
handwritten text recognition across heterogeneous collections. It combines
two components that act at different points of the learning problem. The
first, \emph{structure-aware multitask supervision}, changes what the model
is trained to predict: structural annotations from all sources are mapped to
a common Page--Region--Line representation, from which four complementary
generation tasks are derived. The second, the \emph{pointer-guided
fine-grained Mixture-of-Experts} (PF-MoE), changes what visual evidence the
decoder can access at each generation step: for every generated token, it
retrieves fine multi-scale features of the page at the location being
transcribed and injects them through routed experts. Both components use the
same conditional-generation interface, and neither requires additional
annotation.

%% ------------------------------------------------------------
\subsection{Problem Formulation and Overview}
\label{sec:overview}

Given a document image $I$ and a textual task instruction $\tau$, the
recognizer generates the target sequence $y=(y_1,\ldots,y_T)$ one token at a
time, each token $y_t$ conditioned on $I$, $\tau$, and the previous tokens
$y_{<t}$; we write $p_\theta(y_t\mid y_{<t},I,\tau)$ for this distribution.
Throughout the paper, $t$ indexes target (text) tokens. We build on
Qwen3.5-0.8B-Base, whose visual pathway follows the common encoder--merger
design. The vision encoder splits $I$ into $16\times16$-pixel patches, and a
merger concatenates the features of every $2\times2$ window of neighboring
patches and projects them to the decoder width $d=1024$. Each of the
resulting $N$ visual tokens therefore summarizes a $32\times32$-pixel region
of the page. The visual tokens and the instruction tokens are processed by a
24-layer language decoder, which produces the output text.

\begin{figure*}[tbp]
    \centering
    \includegraphics[width=\textwidth]{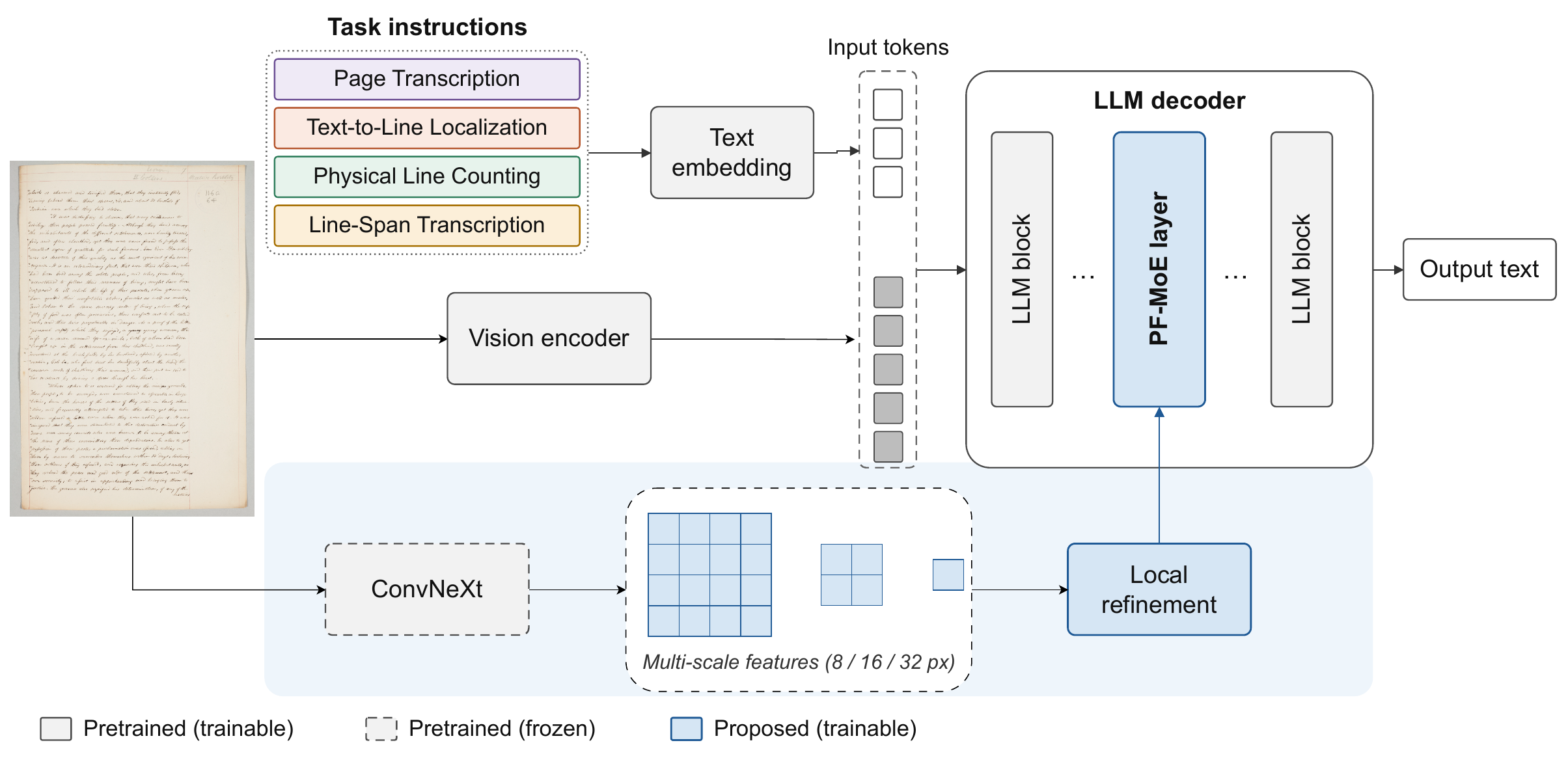}
    \caption{
        Overview of ExpertHTR. A page image and one of four task
        instructions are processed by a vision--language model. The
        pretrained vision encoder, including its $2\times2$ merger, produces
        one visual token per $32\times32$-pixel region. In parallel, a frozen
        ConvNeXt extracts features of the same page at strides of 8, 16, and
        32 pixels, and a local refinement turns them into 21 fine-grained
        cells per visual token. The local refinement is applied to each stride
        separately and consists of a LayerNorm and a linear projection to 256
        channels, followed by a stack of three residual depthwise-separable
        convolution blocks (depthwise convolution, GELU, pointwise
        $1\times1$ convolution): two with $3\times3$ kernels for stroke-level
        context and one with a $1\times9$ kernel for context along the text
        line (Eq.~\eqref{eq:cells}). The cells are read only by the PF-MoE layers
        in the decoder (Fig.~\ref{fig:pfmoe}); the visual token sequence is
        unchanged.
    }
    \label{fig:framework_overview}
\end{figure*}

Figure~\ref{fig:framework_overview} summarizes the framework. During
training, every page contributes records from four tasks
(Sec.~\ref{sec:multitask}). In parallel with the coarse visual tokens, a
fine path encodes the same page into multi-scale cells at strides of 8, 16,
and 32 pixels. Four decoder layers are equipped with PF-MoE
(Sec.~\ref{sec:moe}), which reads these cells for every generated token. At
inference time, only the Page Transcription instruction is used; the
auxiliary tasks therefore add no inference cost.

%% ------------------------------------------------------------
\subsection{Page--Region--Line Representation}
\label{sec:prl}

HTR collections describe document structure with dataset-specific
conventions, such as paragraph boxes, text regions, or line polygons. To
derive supervision consistently across sources, we map every training page to
a common Page--Region--Line representation. A page
$P=(R_1,\ldots,R_M)$ is an ordered sequence of regions, and each region
$R_m=(\ell_{m,1},\ldots,\ell_{m,K_m})$ is an ordered sequence of lines,
where $\ell_{m,k}$ is the transcription of the $k$-th physical
handwritten line of region $R_m$, and the orders of regions and lines follow
the reading order of the source annotation. Line indices restart at one in
every region, and the total number of physical lines is
$L=\sum_{m=1}^{M}K_m$. A region takes the meaning provided by the source
dataset, for example a paragraph or an annotated text block. In our data,
every page of BRESSAY, HWDB2.0, IAM, and ScribbleLens consists of a single
region, whereas pages of Bentham, READ-2016, and RIMES may contain several.

The representation introduces no new labels. Its role is to define a single
coordinate system, the pair $(m,k)$, that addresses every physical line of a
page. The serialization used for transcription,
\begin{equation}
\label{eq:serialization}
S(P)
=
\texttt{[R1]:}\;\ell_{1,1}\,\texttt{\textbackslash n}\,\ell_{1,2}
\,\texttt{\textbackslash n}\cdots\;
\texttt{[R2]:}\;\ell_{2,1}\cdots ,
\end{equation}
marks region boundaries explicitly and encodes physical lines as line breaks.
Because all tasks in Sec.~\ref{sec:multitask} refer to the same $(m,k)$
coordinates, their targets are mutually consistent: the number of line
breaks in $S(P)$, the line count, and the line indices used for localization
and span transcription describe the same physical structure.

%% ------------------------------------------------------------
\subsection{Structure-Aware Multitask Supervision}
\label{sec:multitask}

\begin{figure}[tbp]
    \centering
    \includegraphics[width=\linewidth]{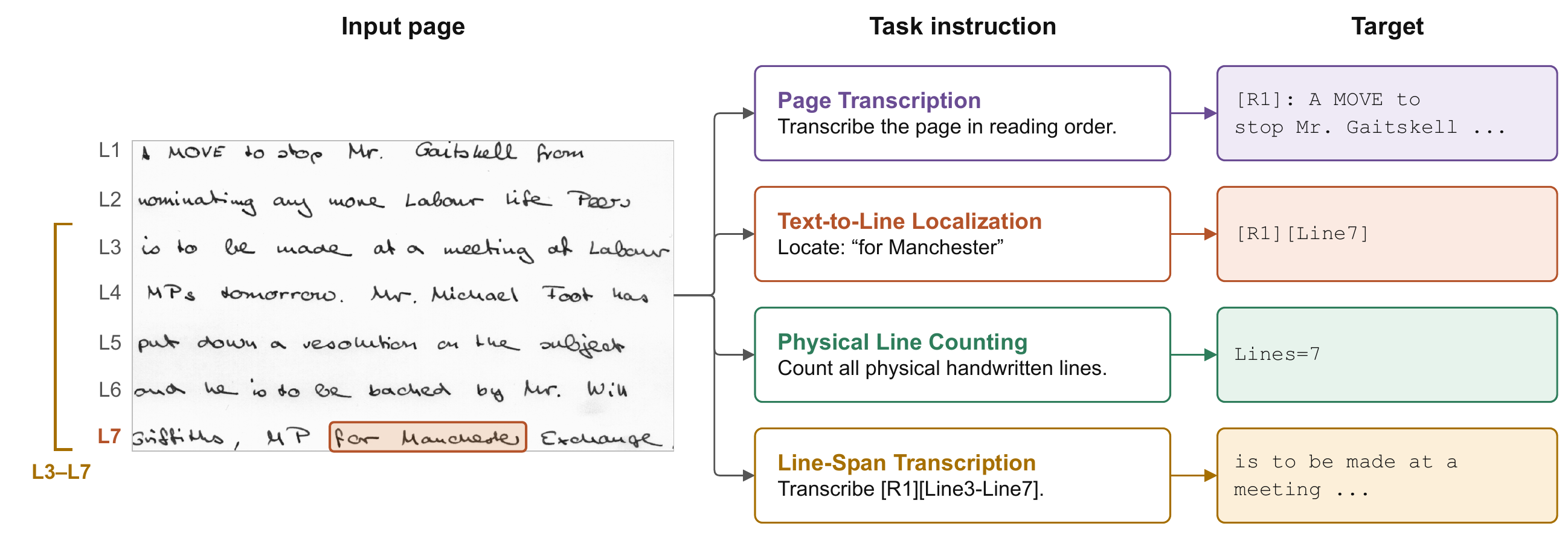}
    \caption{
        The four supervision tasks constructed from the same page. All tasks
        share the input image and the generation interface and differ only in
        the instruction. Page Transcription produces the serialized page with
        region tags; Text-to-Line Localization maps a textual query to its
        region and physical line; Physical Line Counting predicts the number
        of physical lines; Line-Span Transcription transcribes a requested
        line range while the full page remains visible. Outputs are shortened
        for clarity.
    }
    \label{fig:multitask_examples}
\end{figure}

\paragraph{Motivation.} Page-level recognition fails in ways that are rare in
isolated-line recognition. A recognizer may skip or merge physical lines,
re-enter a line it has already transcribed and fall into a repetition loop,
place content in the wrong position of the reading order, or lose local
accuracy as the output sequence grows. When the page transcription is the only
target, all of these failures are penalized through the same token-level
cross-entropy over a sequence that typically contains hundreds of tokens. The
training signal for the underlying structural relations, namely how many lines
exist, where a piece of text is located, and how to transcribe a given part of
the page, is therefore only implicit and is diluted across the long sequence.
We make these relations explicit by deriving auxiliary tasks from $P$ that
isolate each relation in a short target. This follows the classical view of
multitask learning as an inductive bias from related tasks that share a
representation~\citep{caruana1997multitask}, specialized here to the structure
of page-level HTR. Each task is defined by an instruction template and a
target, both computed deterministically from $P$ and, where applicable, from a
sampled argument. Fig.~\ref{fig:multitask_examples} shows the four tasks
constructed from a single page.

% \begin{table}[tbp]
% \caption{
% The four supervision tasks. All tasks take the complete page image as input
% and differ only in the instruction and the target. $(m,k)$ indexes the $k$-th
% physical line of region $R_m$.
% }
% \label{tab:tasks}
% \begin{tabular*}{\linewidth}{@{} LLLL @{}}
% \toprule
% Task & Argument & Target & Supervised relation \\
% \midrule
% Page Transcription (PT)        & --              & $S(P)$                                     & Full transcription, reading order \\
% Physical Line Counting (PLC)   & --              & \texttt{Lines=}$L$                         & Physical-line coverage \\
% Text-to-Line Localization (T2L)& phrase $q$      & \texttt{[R$m$][Line$k$]} or \texttt{0}     & Content--position grounding \\
% Line-Span Transcription (LST)  & span $(m,a,b)$  & $\ell_{m,a},\ldots,\ell_{m,b}$             & Localized recognition \\
% \bottomrule
% \end{tabular*}
% \end{table}

\paragraph{Page Transcription (PT).}
PT is the primary task and the only one used at inference. Given the page
image, the model generates $S(P)$ in Eq.~\eqref{eq:serialization}. Retaining
region tags and line breaks in the target means that the output can be mapped
back to the page structure, and that the transcription shares the coordinate
system of the auxiliary tasks.

\paragraph{Physical Line Counting (PLC).}
PLC asks for the total number of physical handwritten lines across all
regions, with target \texttt{Lines=}$L$. Omitted or duplicated lines in PT are
errors in coverage, but under the PT objective they are visible only through
misaligned character tokens. PLC provides a direct, page-level signal for
coverage in a target of a few tokens. It supervises the count only and does
not require the model to predict geometry.

\paragraph{Text-to-Line Localization (T2L).}
T2L links textual content to its position. In a positive record, a phrase $q$
of consecutive words is taken from one physical line $\ell_{m,k}$ and given in
the instruction, and the target is the identifier \texttt{[R$m$][Line$k$]}.
Phrases contain two to five words; for HWDB2.0, which has no word delimiters,
they contain seven or eight characters. A phrase is used only if it occurs
exactly once on the page, so that every positive target is unambiguous.
Letter case, punctuation, and diacritics are preserved, so the model must
match the query at the character level. Solving T2L requires recognizing the
phrase in the image and resolving its position in the reading order, which is
the relation violated when content is transcribed at the wrong place or a
line is entered twice.

Without further constraints, a model could solve T2L by always returning the
most plausible line, without verifying that the query is actually written on
the page. We therefore include negative records, in which the query does not
occur on the page and the target is \texttt{0}. Negative records make up
10\% of the T2L records and require the model to check the presence of the
queried text, which discourages grounding hallucinated content.

\paragraph{Line-Span Transcription (LST).}
LST asks for the transcription of a contiguous span of lines
$(\ell_{m,a},\ldots,\ell_{m,b})$ within one region, specified in the
instruction as \texttt{[R$m$][Line$a$-Line$b$]}. Unlike conventional
line-level training, the image is not cropped: the full page remains the
visual input, and only the output is restricted. The task therefore trains
the decoder to begin transcription at an arbitrary structural position and to
stop at another, using the same global visual context as PT. This provides
shorter, localized recognition targets and exercises the skills needed to
continue correctly after a line transition.

Spans cover three to seven consecutive lines. Because page-level failures
accumulate on long pages, the number of spans grows with page length: pages
with fewer than 15 physical lines receive one span, pages with 15 to 29 lines
receive two, and pages with 30 or more lines receive three. Spans from the
same page are sampled independently and may overlap.

\paragraph{Common properties of the tasks.}
The four tasks share three properties. First, all targets are deterministic
functions of existing annotations, so no additional labeling is needed.
Second, all tasks use the same instruction-conditioned generation interface,
without task-specific heads or loss terms; the auxiliary tasks influence the
model only through the shared parameters. Third, the auxiliary targets are
short relative to PT, so each auxiliary record concentrates its gradient on a
single structural relation. All records are constructed once before
training and are kept fixed across epochs.

%% ------------------------------------------------------------
\subsection{Pointer-Guided Fine-Grained Mixture-of-Experts}
\label{sec:moe}

\paragraph{Motivation.}
The $2\times2$ merger (Sec.~\ref{sec:overview}) reduces the number of visual
tokens by a factor of four, which keeps page-level decoding tractable, but
each visual token then summarizes a $32\times32$-pixel region. For
handwriting, the discriminative evidence is often much smaller: diacritics,
cancellation strokes, short connecting strokes, and the components of dense
Chinese characters may occupy only a few pixels of such a region. When this
detail is not recoverable from the merged token, the decoder can only rely on context and
its language prior. We therefore hypothesized that access to finer evidence
would help, and we test this hypothesis in Sec.~\ref{sec:analysis}. Two
observations shape how finer evidence should be
provided. First, fine detail is needed only locally, at the position that is
being transcribed; raising the input resolution would instead multiply the
number of visual tokens of every page. Second, the decoder's own attention
already indicates where it is reading, so the location of the evidence can be
derived from the decoder rather than from a separate localization module.

\begin{figure*}[tbp]
    \centering
    \includegraphics[width=0.92\textwidth]{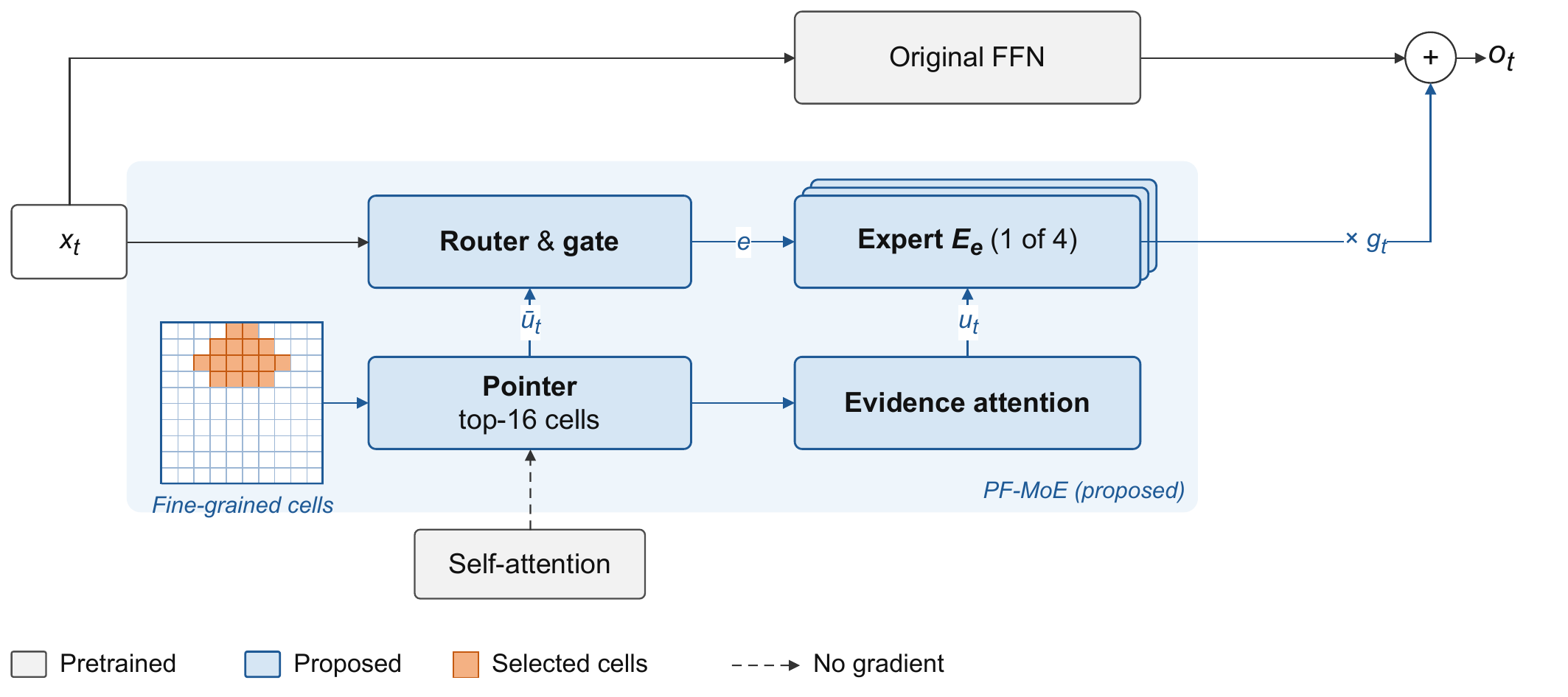}
    \caption{
        One PF-MoE layer for a text token $t$. The original FFN is kept
        (top). The new components (shaded) form two paths: a pointer,
        initialized from the attention of the same layer (dashed: no gradient
        through the prior), selects 16 fine-grained cells of the page (orange),
        summarizes them into $\bar{\mathbf{u}}_t$ for the router and gate, and passes
        them to the evidence attention, which reads them into $\mathbf{u}_t$; a
        top-1 router selects one of four experts, whose output is scaled by
        $g_t$ and added to the FFN output. Visual tokens and tokens before the
        image use only the original FFN. The diagram was prepared with the
        assistance of an AI tool and checked by the authors.
    }
    \label{fig:pfmoe}
\end{figure*}

PF-MoE follows these observations with four design principles:
(i) fine multi-scale features of the page are computed once and kept outside
the token sequence;
(ii) for every generated token, a pointer initialized from the decoder's
attention selects a small set of fine-grained cells;
(iii) the selected evidence enters the hidden state through conditionally
routed experts, so that tokens from different scripts, periods, and writing
styles can be processed by different parameters, while the sequence length is
unchanged; and
(iv) the module is initialized so that the model reproduces the dense model
exactly at the start of training.
Figure~\ref{fig:pfmoe} illustrates one PF-MoE layer, and the components are
described below in the order in which they are computed.

\paragraph{Multi-scale cells.}
The fine path describes each $32\times32$-pixel region covered by a visual
token at three finer resolutions. At stride $\delta\in\{8,16,32\}$ pixels,
the region contains $(32/\delta)^2$ positions, so it is described by
$16$ cells of $8\times8$ pixels, $4$ cells of $16\times16$ pixels, and one
cell of $32\times32$ pixels, that is, $21$ cells per visual token
(Fig.~\ref{fig:framework_overview}). A page with $N$ visual tokens therefore
has $21N$ cells. The cells are not added to the token sequence of the
decoder; they form a separate memory that only the PF-MoE layers read. They
are computed in three steps.

\textit{Backbone features.} The page image that the vision encoder receives,
reconstructed from its input patches, is encoded by an ImageNet-pretrained
ConvNeXt-Tiny \citep{convnext}, which is frozen and kept in evaluation mode.
We take the outputs of its stages at strides 8, 16, and 32, with
$C_\delta=192$, $384$, and $768$ channels, and denote the feature map of
stride $\delta$ by
$F_\delta\in\mathbb{R}^{C_\delta\times H_\delta\times W_\delta}$.

\textit{Local refinement.} The ConvNeXt features are trained for image
classification, not for handwriting. Each map is therefore projected to a
common width $d_f=256$ and refined on its own grid by three residual
depthwise-separable convolution blocks:
\begin{align}
\label{eq:cells}
Z^{(0)}_\delta &= W_\delta\,\operatorname{LN}_\delta(F_\delta), \nonumber\\
Z^{(b)}_\delta &= Z^{(b-1)}_\delta
+ \operatorname{PW}^{(b)}_\delta\Big(\operatorname{GELU}\big(\operatorname{DW}^{(b)}_\delta\big(Z^{(b-1)}_\delta\big)\big)\Big),
\qquad b=1,2,3 .
\end{align}
Here $\operatorname{LN}_\delta$ is a layer normalization over channels and
$W_\delta\in\mathbb{R}^{d_f\times C_\delta}$ a linear projection applied at
every grid position. In block $b$, $\operatorname{DW}^{(b)}_\delta$ is a
depthwise convolution and $\operatorname{PW}^{(b)}_\delta$ a pointwise
($1\times1$) convolution, each with its own parameters. The depthwise kernel
is $3\times3$ in the first two blocks, which combine neighboring stroke
fragments, and $1\times9$ (one row, nine columns) in the third, which adds
context along the text line. The pointwise convolutions are initialized to
zero, so each stride starts as a plain projection. We write
$Z_\delta=Z^{(3)}_\delta$ for the refined map.

\textit{Cell embeddings.} The cell at position
$p\in\{1,\ldots,(32/\delta)^2\}$ of stride $\delta$ inside visual token $i$
is the refined feature at that position plus two learned embeddings,
$\mathbf{z}_{i,\delta,p}=Z_\delta[i,p]+\mathbf{e}^{\mathrm{str}}_{\delta}
+\mathbf{e}^{\mathrm{pos}}_{\delta,p}\in\mathbb{R}^{d_f}$, which tell the
later modules the stride of the cell and its position inside the token. In
the following, we write $\mathbf{z}_c$ for a cell $c$, $i(c)$ for the visual
token that contains it, and $\mathcal{Z}$ for the set of all $21N$ cells of
the page. The cells are computed once per page and shared by all PF-MoE
layers and decoding steps.

\paragraph{Attention-guided pointer.}
For each generated text token $t$, the pointer chooses which of the $21N$
cells the layer reads. It works from coarse to fine: the attention of the decoder indicates which
visual tokens are being read, and a learned score then chooses among their
cells. The coarse part is the \emph{attention prior} $a_{t,i}$: the
softmax attention weight of token $t$ on visual token $i$ in the same layer,
restricted to the $N$ visual tokens and averaged over the attention heads. It
is computed with the layer's own query and key projections, query--key
normalization, and rotary position encoding, and without gradient. Let
$\tilde{\mathbf{h}}_t\in\mathbb{R}^{d}$ be the normalized input of the
attention sublayer for token $t$. For a cell $c$, the pointer score is
\begin{equation}
\label{eq:pointer_score}
s_{t,c}
=
\log a_{t,i(c)}
+
\frac{\left(P_q\,\tilde{\mathbf{h}}_t\right)^{\!\top}\!\left(P_k\,\mathbf{z}_c\right)}{\sqrt{d_p}},
\end{equation}
where $P_q\in\mathbb{R}^{d_p\times d}$ and $P_k\in\mathbb{R}^{d_p\times d_f}$
are the learned query and key projections of the pointer and $d_p=128$. All
cells of the same visual token share the prior term. The first term places the
pointer where the decoder reads; the learned second term chooses among the
cells of those visual tokens, and thereby among the three strides, and may
move the pointer away from the attention maximum. 

The pointer selects the $n_c=16$ cells with the largest scores among all cells of the page,
$\mathcal{C}_t=\operatorname{TopK}_{n_c}\{s_{t,c}:c\in\mathcal{Z}\}$, and
normalizes their scores with a softmax over $\mathcal{C}_t$ to obtain the
pointer weights $w_{t,c}$. Because the selection runs over all cells of the
page, the selected cells need not belong to a single visual token; they can
come from several neighboring tokens and mix the three strides. The selection
itself is discrete, but the weights carry gradient into $P_q$
and $P_k$ through the modules that consume them. During training, standard
Gumbel noise is added to the scores used for selection only, which lets the
pointer explore cells outside its current top-$n_c$. $P_q$ is initialized
close to zero, so that at the start of training the choice follows the
decoder's attention. The selected cells are read only by the evidence
attention described next and are never appended to the token sequence.

\paragraph{Evidence attention.}
Let $\mathbf{x}_t\in\mathbb{R}^{d}$ denote the normalized input of the FFN
sublayer for token $t$, that is, the input that the original FFN receives.
The selected cells are read by a small cross-attention with $H_f=4$ heads of
width $d_f/H_f=64$. Its query depends on $\mathbf{x}_t$ and on the expert
$e_t$ chosen by the router (Eq.~\eqref{eq:router}), so that each expert learns
what to look for: $\bar{\mathbf{q}}_t=Q_{e_t}\mathbf{x}_t$ with one
$Q_e\in\mathbb{R}^{d_f\times d}$ per expert. Keys and values are linear
projections of the cells, $\bar{\mathbf{k}}_c=W_K\mathbf{z}_c$ and
$\bar{\mathbf{v}}_c=W_V\mathbf{z}_c$ with $W_K,W_V\in\mathbb{R}^{d_f\times d_f}$,
and the pointer weights are added to the attention logits as a log-prior.
In each head,
\begin{equation}
\label{eq:fine_look}
\alpha_{t,c}
=
\operatorname*{softmax}_{c\in\mathcal{C}_t}
\left(\frac{\bar{\mathbf{q}}_t^{\top}\bar{\mathbf{k}}_c}{\sqrt{d_f/H_f}}+\log w_{t,c}\right),
\end{equation}
computed on the $d_f/H_f$-dimensional slices of the query and keys of that
head. Each head returns the $\alpha$-weighted sum of its values
$\bar{\mathbf{v}}_c$; the outputs of the four heads are concatenated and projected
by $W_O\in\mathbb{R}^{d\times d_f}$ to the evidence vector
$\mathbf{u}_t\in\mathbb{R}^{d}$. Keys and values depend only on the page and
the layer and are projected once per page.

\paragraph{Evidence-conditioned routing and gating.}
Routing and gating are conditioned on the current state and on a summary of
the selected evidence,
$\bar{\mathbf{u}}_t=\sum_{c\in\mathcal{C}_t}w_{t,c}\,\mathbf{z}_c\in\mathbb{R}^{d_f}$.
A router with weights $W_r\in\mathbb{R}^{N_e\times(d+d_f)}$ selects one of
$N_e=4$ experts,
\begin{equation}
\label{eq:router}
\mathbf{p}_t=\operatorname{softmax}\!\left(W_r\,[\mathbf{x}_t;\bar{\mathbf{u}}_t]\right)\in\mathbb{R}^{N_e},
\qquad
e_t=\arg\max_{e\in\{1,\ldots,N_e\}}\,p_{t,e},
\end{equation}
and a correction gate with weights $\mathbf{w}_g\in\mathbb{R}^{d+d_f}$ and
scalar bias $b_g$ decides how strongly the routed update is applied,
\begin{equation}
\label{eq:gate}
\gamma_t=\operatorname{sigmoid}\!\left(\mathbf{w}_g^{\top}[\mathbf{x}_t;\bar{\mathbf{u}}_t]+b_g\right),
\qquad
g_t=p_{t,e_t}\,\gamma_t ,
\end{equation}
where $p_{t,e_t}$ is the router probability of the selected expert. The gate
weights $\mathbf{w}_g$ are initialized to zero and the bias to $b_g=2$, so the
gate starts open with $\gamma_t=\operatorname{sigmoid}(2)\approx0.88$ for
every token and learns where the update should be reduced. Routing never uses language,
script, dataset, or task identities.

\paragraph{Evidence-augmented experts.}
Each expert $E_e$, $e\in\{1,\ldots,N_e\}$, has the same form as the FFN of
the backbone, a bias-free SwiGLU MLP with gate, up, and down projections,
but its input is the concatenation of the linguistic state and the retrieved
evidence, $\mathbf{r}_t=[\mathbf{x}_t;\mathbf{u}_t]\in\mathbb{R}^{2d}$. Its
hidden width is $d_e=896$, so the four experts together are as wide as the
original FFN ($4\times896=3584$), and its down projection
$W^{e}_{\mathrm{down}}\in\mathbb{R}^{d\times d_e}$ maps back to the decoder
width. Let
$D_l$ denote the original FFN of layer $l$, which is kept and trained. For
text tokens that follow the image, the FFN sublayer of layer $l$ outputs
\begin{equation}
\label{eq:pfmoe}
\mathbf{o}_t
=
D_l(\mathbf{x}_t)
+
g_t\,E_{e_t}(\mathbf{r}_t),
\end{equation}
which is added to the residual stream as usual. Visual tokens and the text
tokens that precede the image use only $D_l$. Because decoding is causal,
the representations of these tokens, and hence the visual keys from which the
pointer prior is computed, are produced by the backbone alone.

\paragraph{Layer placement.}
The decoder of Qwen3.5 interleaves Gated DeltaNet linear-attention layers
with full softmax-attention layers; every fourth layer (layers 3, 7, 11, 15,
19, and 23, indexed from zero) uses full attention. The pointer prior in
Eq.~\eqref{eq:pointer_score} requires explicit attention scores between a
text token and individual visual tokens, which are available only in
full-attention layers. We therefore place PF-MoE in four of them,
$\mathcal{L}_{\mathrm{MoE}}=\{7,11,15,19\}$, and keep all other FFNs
unchanged. PF-MoE adds 108.6M parameters to the
852.74M parameters of the backbone: 27.8M in the frozen ConvNeXt stages and
80.8M trained parameters.

\paragraph{Function-preserving initialization.}
PF-MoE is inserted into the dense multitask checkpoint obtained after the
first training stage (Sec.~\ref{sec:training_objective}). The expert output projections
$W^{e}_{\mathrm{down}}$ are initialized to zero, while all other expert,
router, and evidence-attention weights are initialized randomly. The routed term in
Eq.~\eqref{eq:pfmoe} therefore vanishes, and the model reproduces the dense
model exactly at the first step, which we verified numerically. Because the
input projections of the experts are non-zero, their hidden activations are
non-zero from the start, so $W^{e}_{\mathrm{down}}$ receives gradient at the
first update and the gradient then reaches the evidence attention, the pointer, and
the local refinement.

\paragraph{Inference cost.}
The fine-grained cells, the evidence-attention keys and values, and the pointer keys
$P_k\mathbf{z}_c$ of each PF-MoE layer are computed once during the
prefill pass. At each decoding step, PF-MoE computes the pointer and the
evidence attention for the new token only, over $n_c=16$ cells. The visual sequence
length remains $N$ throughout, whereas doubling the input resolution along
each axis would increase it to $4N$ for every page, with a corresponding
increase in attention cost and key--value memory.

%% ------------------------------------------------------------
\subsection{Training Objective and Schedule}
\label{sec:training_objective}

\paragraph{Language-modeling loss.}
All four tasks are optimized with the same autoregressive objective. The loss
is computed only on target tokens; instruction and image tokens are masked.
Page transcription is prone to termination errors: the model may stop before
the last line or continue past it in a repetition loop. We therefore increase
the weight of the end-of-sequence token in PT targets:
\begin{equation}
\label{eq:lm_loss}
\mathcal{L}_{\mathrm{LM}}
=
-\frac{1}{\sum_{t\in\mathcal{A}}\beta_t}
\sum_{t\in\mathcal{A}}
\beta_t\log p_\theta\!\left(y_t\mid y_{<t},I,\tau\right),
\qquad
\beta_t=
\begin{cases}
\omega, & \text{$y_t$ is the end token of a PT target},\\
1, & \text{otherwise},
\end{cases}
\end{equation}
where $\mathcal{A}$ is the set of target positions in the effective batch,
$I$ and $\tau$ are the image and instruction of the record that contains
position $t$, and $\omega=4$. The loss is thus a weighted mean over target
tokens rather than over sequences.

\paragraph{Load balancing.}
To prevent the router from collapsing onto a single expert, we use the
load-balancing loss of Switch Transformer~\citep{switch_transformer}. At each
PF-MoE layer, it is computed over the routed tokens of the batch as
$N_e\sum_{e=1}^{N_e}f_e P_e$, where $f_e$ is the fraction of tokens routed to
expert $e$ and $P_e$ the mean router probability $p_{t,e}$; it equals 1
when routing is uniform. $\mathcal{L}_{\mathrm{bal}}$ is the mean of this
loss over the four PF-MoE layers.

\paragraph{Two-stage training.}
Training proceeds in two stages that use the same multitask data. In the
first stage, all parameters of the pretrained VLM are fine-tuned with
$\mathcal{L}_{\mathrm{LM}}$ on the four tasks for three epochs. In the second
stage, PF-MoE is inserted into this checkpoint as described above, and the
full model is trained for one epoch with
$\mathcal{L}=\mathcal{L}_{\mathrm{LM}}+\lambda_{\mathrm{bal}}\mathcal{L}_{\mathrm{bal}}$,
where $\lambda_{\mathrm{bal}}$ is increased linearly from 0 to 0.01 over the first
10\% of the updates of the training epoch. $\mathcal{L}_{\mathrm{bal}}$ is computed
for each micro-batch and weighted by its number of target tokens when the
micro-batches of an update are combined. The ConvNeXt remains frozen; all other parameters, including
the original FFNs, are updated. The new modules start from zero contribution
while the backbone is already adapted to HTR. They are therefore trained with
a learning rate 25 times larger than that of the backbone
(Sec.~\ref{sec:experimental_setup}).

\section{Experiments}
\label{sec:experiments}
This section evaluates ExpertHTR on the seven handwriting datasets used in
this study. After the experimental setup (Sec.~\ref{sec:experimental_setup}),
we compare ExpertHTR with specialized systems and with zero-shot and
fine-tuned OCR/VLM models (Sec.~\ref{sec:main_results}), isolate the
contribution of the multitask training data (Sec.~\ref{sec:data_effect}),
and test the transfer of PF-MoE to other backbones
(Sec.~\ref{sec:backbones}). Ablations of the tasks and of the components of
PF-MoE follow in Sec.~\ref{sec:ablation}, and Secs.~\ref{sec:analysis}
and~\ref{sec:error_analysis} analyze what PF-MoE learns and which errors
remain.

\subsection{Experimental Setup}
\label{sec:experimental_setup}

\paragraph{Datasets.}
We evaluate ExpertHTR on seven HTR datasets: BRESSAY, Bentham R0,
HWDB2.0, IAM, READ-2016, RIMES, and ScribbleLens. Together, these
collections cover multiple languages and scripts, as well as modern and
historical handwriting with different document characteristics.
Table~\ref{tab:datasets} summarizes the training, validation, and test
partitions used in our experiments.

For BRESSAY, IAM, READ-2016, and ScribbleLens, we follow the evaluation
configurations used by Meta-DAN~\citep{metadan_htr} to maintain comparability
with previously reported document-level HTR results. IAM is evaluated using
the corresponding paragraph-level configuration. Bentham R0 is included as
an additional historical benchmark that is not part of the Meta-DAN
evaluation suite.

For Chinese handwriting, we use the HWDB2.0 subset of the CASIA-HWDB
database~\citep{casia_hwdb}. The handwritten-text portion of CASIA-HWDB
contains HWDB2.0, HWDB2.1, and HWDB2.2; only HWDB2.0 is used in this work,
and its official HWDB2.0Test partition is retained for evaluation. We
therefore refer to this benchmark consistently as \textit{HWDB2.0} rather
than using the broader CASIA2 designation adopted in some previous
implementations.

For RIMES, we use correspondence-letter documents (\texttt{\_L}) from RIMES
Complete~\citep{rimes_complete}. The release also contains questionnaires
(\texttt{\_Q}) and faxes (\texttt{\_F}), whose document structure and
printed/form content differ from the handwritten correspondence setting
considered here. We therefore restrict training to the letter subset.
Because RIMES is considerably larger than the other training sources, we
retain 758 of the 5,054 available \texttt{\_L} training pages, approximately
15\% of the original training partition, to limit its dominance in the joint
training mixture. The validation and test partitions are not subsampled.

\begin{table}[tbp]
\centering
\caption{
Datasets used in the main experiments. Training-set curation does not modify
the corresponding validation or test partitions. $^{\diamond}$\,RIMES: only the
handwritten correspondence letters (\texttt{\_L} pages) of RIMES Complete are
used in all three partitions; questionnaires (\texttt{\_Q}) and faxes
(\texttt{\_F}) are excluded. The test partition contains all 273 letter pages
of the test split.
}
\label{tab:datasets}

\footnotesize
\setlength{\tabcolsep}{4pt}
\renewcommand{\arraystretch}{1.05}

\begin{tabular*}{\linewidth}{@{\extracolsep{\fill}} llrrr @{}}
\toprule
Dataset & Language & Train & Validation & Test \\
\midrule
BRESSAY      & Portuguese & 647  & 154 & 199 \\
Bentham R0   & English    & 350  & 50  & 33  \\
HWDB2.0      & Chinese    & 1377 & 300 & 415 \\
IAM          & English    & 747  & 116 & 336 \\
READ-2016    & German     & 350  & 50  & 50  \\
RIMES$^{\diamond}$      & French     & 758  & 278 & 273 \\
ScribbleLens & Dutch      & 164  & 14  & 21  \\
\midrule
Total        & --         & 4393 & 962 & 1327 \\
\bottomrule
\end{tabular*}
\end{table}

\paragraph{Multitask data construction.}
The annotations from the seven training sources are converted to the
Page--Region--Line representation described in Sec.~\ref{sec:prl}. In the curated training set used in our
experiments, the available physical-line information is sufficient to
construct Page Transcription, Physical Line Counting, and Text-to-Line
Localization for every training image. Line-Span Transcription produces
additional records from contiguous ranges of annotated physical lines within
the same region.

Each image contributes one Page Transcription record, one Physical Line
Counting record, and one Text-to-Line Localization record. Depending on the
available line structure, one to three Line-Span Transcription records are
also generated for each image. The 4,393 training images produce 20,388
multitask records in total, as summarized in
Table~\ref{tab:task_distribution}. Records are shuffled without replacement
and visited once per epoch, with the curated source frequencies described
above and without source-balanced sampling or oversampling. The experiments
therefore evaluate the benefit of complementary supervision constructed from
the annotations available in the training data, rather than a controlled
setting in which particular annotation types are intentionally removed from
selected datasets.

\begin{table}[tbp]
\centering
\caption{
Distribution of training records across the four multitask supervision tasks.
}
\label{tab:task_distribution}

\footnotesize
\setlength{\tabcolsep}{4pt}
\renewcommand{\arraystretch}{1.05}

\begin{tabular*}{.78\linewidth}{@{\extracolsep{\fill}} lrr @{}}
\toprule
Task & Records & Share (\%) \\
\midrule
Page Transcription          & 4393 & 21.55 \\
Physical Line Counting      & 4393 & 21.55 \\
Text-to-Line Localization   & 4393 & 21.55 \\
Line-Span Transcription     & 7209 & 35.36 \\
\midrule
Total                       & 20388 & 100.00 \\
\bottomrule
\end{tabular*}
\end{table}

\paragraph{Evaluation metrics.}
We evaluate recognition performance using character error rate (CER) as the
primary metric and word error rate (WER) as a complementary metric. Both
metrics are computed using Levenshtein edit distance and are micro-aggregated
over the test set. WER is not reported for HWDB2.0 because we do not
introduce an additional Chinese word-segmentation procedure. In the ablation
study we also report the macro-averaged CER, the unweighted mean of the
seven per-dataset CERs, and the number of added parameters.
For the component ablation and the analyses in Secs.~\ref{sec:analysis}
and~\ref{sec:error_analysis}, we additionally match every reference line to its closest predicted line and
count skipped and repeated lines per 100 reference lines, count the pages for
which generation reaches the token limit, and estimate 95\% intervals with a
paired bootstrap over test pages.

\paragraph{Implementation details.}
The main model is built on Qwen3.5-0.8B-Base and is the only model trained
with full fine-tuning; all other fine-tuned models (the larger backbones of
Sec.~\ref{sec:backbones} and the OCR/VLM baselines of
Tables~\ref{tab:main_results} and~\ref{tab:data_effect}) use the LoRA setting
described below. In the first stage, the dense multitask model is trained for
three epochs with AdamW (learning rate $4\times10^{-6}$, effective batch size
32). In the second stage, PF-MoE is inserted and the model is trained for one
more epoch, with a learning rate of $1\times10^{-4}$ for the new modules and
$4\times10^{-6}$ for the backbone; the ConvNeXt stays frozen. As the baseline,
the same three-epoch checkpoint is trained for one more epoch without PF-MoE,
so that both models receive the same number of updates. We denote this model
as \emph{Dense} throughout the experiments.

Input images are resized with their aspect ratio preserved, to an area between
0.59 and 1.57 megapixels, and no data augmentation is applied. Decoding is
greedy with at most 1,024 new tokens. All models are trained on a single
NVIDIA RTX PRO 6000 GPU (96\,GB); the remaining settings are given in the
released code.

\paragraph{LoRA fine-tuning.}
All LoRA-adapted models use the same configuration, whether trained per
dataset or on all datasets: rank 64, $\alpha=128$, and dropout 0.05 on the
attention and MLP projections, trained for one epoch with AdamW (learning
rate $1\times10^{-4}$, effective batch size 32) and then merged into the
model. The LoRA branches cover the language model of Qwen3.5-2B (43.6M
trainable parameters) and, for churro-3B, also the MLPs of the vision
encoder (148.6M). For the PF-MoE experiments of Sec.~\ref{sec:backbones}, PF-MoE is
inserted into the best multi-source Qwen3.5-2B and churro-3B LoRA
checkpoints and trained together with new LoRA branches of the same
configuration, while the merged weights stay frozen. For churro-3B, the
image size and generation length follow its original settings (see the
released code).

\subsection{Main Results}
\label{sec:main_results}

Table~\ref{tab:main_results} places ExpertHTR among systems that differ in
how much handwriting-specific training they receive. The specialized
DAN-family systems train one compact model per dataset with synthetic
pretraining matched to the target collection
\citep{dan_htr,fasterdan_htr,metadan_htr}. Released OCR/VLM checkpoints are
evaluated zero-shot. To separate the effect of the training data from that
of the model, we also fine-tune the same VLMs with LoRA on our training
data, either separately for each dataset or jointly on all seven. ExpertHTR uses a single
checkpoint for all seven collections and no synthetic data. All rows except
the published specialized results are evaluated with the same protocol on
our test sets.

\begin{table*}[tbp]
\centering
\caption{
CER and WER (\%) on the seven test sets (no WER for HWDB2.0). $\dagger$:
results of DAN, Faster DAN, and Meta-DAN as reported by
\citet{metadan_htr}, who train all three in the same framework, one model per
dataset, and Bentham is not reported. Their RIMES and HWDB2.0 values are
our evaluation of the released checkpoints on our test sets: for RIMES,
because we use only the 273 handwritten letter pages (\texttt{\_L}) of the
test set, whereas \citet{metadan_htr} report the full RIMES test set; for
HWDB2.0, because they report CASIA-2 instead. All other rows are
evaluated with the same protocol on our test sets. \textit{Param.} is the number of parameters at inference in
millions (LoRA merged; for ExpertHTR including the frozen ConvNeXt stages).
Dense is the same Qwen3.5-0.8B multitask model without PF-MoE
(Sec.~\ref{sec:experimental_setup}). Qwen3.5-0.8B (Dense and ExpertHTR) is the only
fully fine-tuned model; all other fine-tuned models use LoRA. The best and second-best results in each
column are shown in bold and underlined.
}
\label{tab:main_results}

\scriptsize
\setlength{\tabcolsep}{1.6pt}
\renewcommand{\arraystretch}{1.08}

\begin{tabular*}{\textwidth}
{@{\extracolsep{\fill}} l r rr rr r rr rr rr rr @{}}
\toprule
& & \multicolumn{2}{c}{BRESSAY} & \multicolumn{2}{c}{Bentham} & HWDB2.0
& \multicolumn{2}{c}{IAM} & \multicolumn{2}{c}{READ-2016}
& \multicolumn{2}{c}{RIMES} & \multicolumn{2}{c}{ScribbleLens} \\
\cmidrule(lr){3-4}\cmidrule(lr){5-6}\cmidrule(lr){7-7}\cmidrule(lr){8-9}
\cmidrule(lr){10-11}\cmidrule(lr){12-13}\cmidrule(l){14-15}
Method & Param. & CER & WER & CER & WER & CER & CER & WER & CER & WER & CER & WER & CER & WER \\
\midrule
\multicolumn{15}{@{}l}{\textit{Specialized HTR, one model per dataset, synthetic pretraining$^\dagger$}} \\
DAN & 23.1 & 2.43 & \textbf{7.85} & -- & -- & \textbf{0.84} & 4.23 & 14.00 & \textbf{3.85} & \textbf{14.69} & 3.72 & \underline{8.53} & \underline{4.38} & \underline{19.66} \\
Faster DAN & 23.1 & \textbf{2.34} & \underline{8.03} & -- & -- & 1.40 & 4.16 & 13.42 & 4.34 & \underline{14.83} & 3.45 & \textbf{8.02} & 4.81 & \textbf{18.64} \\
Meta-DAN & 23.4 & \underline{2.38} & 8.46 & -- & -- & 1.59 & 4.79 & 15.78 & \underline{4.22} & 16.03 & \textbf{3.25} & 8.72 & \textbf{4.31} & 19.71 \\
\midrule
\multicolumn{15}{@{}l}{\textit{VLMs, zero-shot}} \\
PaddleOCR-VL & 900 & 58.58 & 108.00 & 9.80 & 18.30 & 6.52 & 6.25 & 12.00 & 381.76 & 399.08 & 17.48 & 49.97 & 206.92 & 275.15 \\
DeepSeek-OCR & 3336 & 174.72 & 280.22 & 35.95 & 75.81 & 22.03 & 9.54 & 18.20 & 1502.88 & 2472.80 & 23.44 & 23.92 & 269.22 & 516.61 \\
churro-3B & 3755 & 40.60 & 55.78 & 27.17 & 34.62 & 147.03 & 7.80 & 9.49 & 156.41 & 237.52 & 27.77 & 34.26 & 49.28 & 93.48 \\
Qwen3.5-0.8B & 852.7 & 51.94 & 74.67 & 27.13 & 36.97 & 25.43 & 11.55 & 18.87 & 124.49 & 127.96 & 31.66 & 45.22 & 91.45 & 118.32 \\
\midrule
\multicolumn{15}{@{}l}{\textit{VLMs fine-tuned on our data with LoRA, one model per dataset}} \\
Qwen3.5-2B & 2213 & 37.45 & 53.94 & 8.02 & 18.95 & 7.45 & 5.13 & 13.26 & 114.44 & 159.67 & 10.97 & 25.83 & 71.94 & 112.51 \\
churro-3B & 3755 & 22.17 & 33.42 & 5.48 & 11.85 & 1.65 & 2.00 & 4.74 & 26.81 & 60.50 & 5.47 & 10.22 & 17.15 & 48.31 \\
PaddleOCR-VL & 900 & 50.98 & 65.32 & 21.51 & 37.66 & \underline{1.05} & 4.86 & 11.30 & 63.72 & 93.79 & 10.18 & 21.48 & 38.52 & 75.13 \\
DeepSeek-OCR & 3336 & 175.16 & 257.64 & 51.40 & 76.82 & 16.42 & 15.56 & 21.27 & 393.74 & 501.25 & 20.60 & 29.01 & 397.73 & 620.85 \\
\midrule
\multicolumn{15}{@{}l}{\textit{VLMs fine-tuned on our data, one model for all datasets (LoRA unless stated)}} \\
PaddleOCR-VL & 900 & 49.37 & 59.35 & 14.50 & 26.99 & 5.59 & 2.67 & 6.64 & 33.34 & 62.23 & 8.71 & 16.88 & 43.34 & 75.01 \\
DeepSeek-OCR & 3336 & 90.82 & 125.50 & 66.59 & 72.23 & 36.20 & 8.06 & 13.31 & 231.82 & 245.29 & 23.19 & 28.18 & 292.19 & 478.56 \\
Qwen3.5-0.8B (Dense, full) & 852.7 & 11.16 & 20.55 & 15.40 & 28.02 & 2.36 & 2.99 & 7.43 & 22.67 & 53.98 & 4.02 & 10.92 & 22.38 & 51.58 \\
Qwen3.5-2B (LoRA) & 2213 & 30.99 & 45.77 & 8.99 & 18.50 & 3.24 & 3.18 & 8.00 & 48.07 & 96.15 & 3.42 & 8.93 & 39.94 & 77.02 \\
churro-3B (LoRA) & 3755 & 16.83 & 29.50 & \underline{3.18} & \underline{8.05} & 1.93 & \textbf{1.59} & \textbf{4.26} & 10.81 & 32.87 & \textbf{3.25} & 8.98 & 18.94 & 59.78 \\
\midrule
\multicolumn{15}{@{}l}{\textit{ExpertHTR (this work), one model for all datasets}} \\
\textbf{ExpertHTR} (Qwen3.5-0.8B) & 961.4 & 8.17 & 16.75 & 13.49 & 14.89 & 2.50 & 2.96 & 7.66 & 14.35 & 40.38 & 4.05 & 10.95 & 15.18 & 42.07 \\
\textbf{ExpertHTR} (Qwen3.5-2B) & 2506 & 28.96 & 44.77 & 4.55 & 10.67 & 2.92 & 3.43 & 9.55 & 34.22 & 68.65 & \underline{3.35} & 9.81 & 33.56 & 62.62 \\
\textbf{ExpertHTR} (churro-3B) & 4247 & 13.14 & 23.15 & \textbf{2.79} & \textbf{7.61} & 1.55 & \underline{1.62} & \underline{4.30} & 9.28 & 29.63 & 4.48 & 10.09 & 10.80 & 33.93 \\
\bottomrule
\end{tabular*}
\end{table*}

\paragraph{Fine-tuned OCR/VLM models.}
Fine-tuning on our multitask data turns general OCR/VLM checkpoints into
competitive handwriting recognizers. Zero-shot, every VLM exceeds 100\% CER
on at least one collection because of long repetitions or untranscribed
output; after fine-tuning, several of them reach the level of the
specialized systems on individual collections. The jointly fine-tuned
churro-3B, for example, reaches 1.59\% CER on IAM, 3.18\% on Bentham, and
3.25\% on RIMES, the lowest RIMES CER in Table~\ref{tab:main_results}
together with Meta-DAN. These models use no synthetic pretraining and, in
the multi-source setting, a single checkpoint for all collections.
Section~\ref{sec:data_effect} compares this training data with fine-tuning
on the original annotations.

\paragraph{ExpertHTR among VLMs.}
Among the models that use one checkpoint for all collections, ExpertHTR has
the lowest CER on five of the seven collections. On IAM and RIMES, the
jointly fine-tuned churro-3B without PF-MoE is marginally better (1.59\%
versus 1.62\%, and 3.25\% versus 3.35\%), so the best VLM results on these
two collections also come from our training data. With Qwen3.5-0.8B,
ExpertHTR reduces the micro-averaged CER of Dense from 8.81\% to 6.81\% and
obtains the lowest CER and WER on BRESSAY of all evaluated VLMs (8.17\% and
16.75\%). With churro-3B, it obtains the lowest Bentham CER and WER of all
methods (2.79\% and 7.61\%), the lowest CER of all VLMs on READ-2016
(9.28\%) and ScribbleLens (10.80\%), and the best HWDB2.0 CER among the
multi-source models (1.55\%).

\paragraph{Where PF-MoE helps.}
Comparing each backbone with and without PF-MoE on the same multitask data
shows where the fine-grained evidence pays off. On the four collections that
are hardest to read, the historical Bentham, READ-2016, and ScribbleLens and
the BRESSAY essays, PF-MoE lowers CER on all three backbones (12 of 12
backbone--collection pairs), whereas on IAM, RIMES, and HWDB2.0 the changes
are smaller and of either sign. Section~\ref{sec:backbones} reports the
results per backbone, and Sec.~\ref{sec:analysis} examines where these gains
come from.

\paragraph{Comparison with specialized systems.}
DAN, Faster DAN, and Meta-DAN are compact recognizers of about 23M
parameters, each trained for a single collection. Their training is tailored
to that collection in advance: the encoder is pretrained on several million
synthetic text lines rendered from the text of the target training set, and
a curriculum over synthetic pages precedes training on the real pages
\citep{dan_htr,fasterdan_htr,metadan_htr}. Each model therefore sees the
vocabulary, language, and layout of its target collection many times before
it is evaluated, which is a strong advantage on small collections. ExpertHTR
uses neither: one checkpoint serves all seven collections, and no synthetic
data are used. Under these conditions, the specialized systems remain
clearly more accurate on BRESSAY, READ-2016, and ScribbleLens, where the best
of them reaches 2.34\%, 3.85\%, and 4.31\% CER, compared with 8.17\%,
9.28\%, and 10.80\% for ExpertHTR.

On the other collections, ExpertHTR is better or close. On IAM, it reaches
2.96\% CER with Qwen3.5-0.8B and 1.62\% with churro-3B, compared with 4.16\%
for the best specialized system (Faster DAN); most adapted VLMs in
Table~\ref{tab:main_results} share this advantage, which reflects their
large-scale pretraining on modern English text. On HWDB2.0, ExpertHTR with
churro-3B (1.55\%) is more accurate than Meta-DAN (1.59\%) and close to
Faster DAN (1.40\%), although HWDB2.0 is the only Chinese collection in our
training mixture; only DAN (0.84\%) is clearly ahead. On RIMES, ExpertHTR
with Qwen3.5-2B (3.35\%) is more accurate than DAN (3.72\%) and Faster DAN
(3.45\%) and within 0.1 points of Meta-DAN (3.25\%). On Bentham, for which no
specialized result is reported, it obtains the lowest CER of all evaluated
systems. A single model without target-specific synthetic data is thus
better than or close to the specialized systems on IAM, HWDB2.0, and RIMES,
and narrows, but does not close, the gap on the three collections where
target-matched pretraining matters most.

\paragraph{Models for a single collection.}
The multitask data is not tied to the single-checkpoint setting. Fine-tuned
on one collection alone, without synthetic data or PF-MoE, churro-3B reaches
2.00\% CER on IAM, about half that of the best specialized system, and
PaddleOCR-VL reaches 1.05\% on HWDB2.0 (Sec.~\ref{sec:data_effect}). The
recipe can therefore be used either to build one model for many collections,
as ExpertHTR does, or to adapt a smaller model to a specific collection.

\subsection{Effect of the Training Data}
\label{sec:data_effect}

The fine-tuned VLMs in Table~\ref{tab:main_results} are trained on our
multitask data. To measure what this data contributes independently of
PF-MoE, we compare each model fine-tuned with LoRA on our data with the same model
fine-tuned with LoRA on the original annotations of the datasets, which provide only
the full-page transcription as target, without the Page--Region--Line
structure and without the three auxiliary tasks. Both versions use the same
model and the same evaluation protocol. Table~\ref{tab:data_effect}
reports the pairs for which both versions are available.

\begin{table*}[tbp]
\centering
\caption{
CER (\%) of the same models fine-tuned with LoRA on two kinds of training
targets and evaluated with the same protocol. \textit{Page only}: the original
annotations of each dataset, with the full-page transcription as the only
target. \textit{Multitask (ours)}: the Page--Region--Line representation with
the four tasks of Sec.~\ref{sec:multitask}. Bold: lower CER than
\textit{Page only} for the same model and dataset; the last column counts
these datasets.
}
\label{tab:data_effect}

\footnotesize
\setlength{\tabcolsep}{3pt}
\renewcommand{\arraystretch}{1.05}

\begin{tabular*}{\textwidth}
{@{\extracolsep{\fill}} l l rrrrrrr c @{}}
\toprule
Model & Training targets & BRES. & Bent. & HWDB & IAM & READ & RIMES & Scrib. & Improved \\
\midrule
\multicolumn{10}{@{}l}{\textit{Single-source fine-tuning}} \\
Qwen3.5-2B & Page only & 42.62 & 11.45 & 6.56 & 2.88 & 317.80 & 10.24 & 89.48 & \\
 & Multitask (ours) & \textbf{37.45} & \textbf{8.02} & 7.45 & 5.13 & \textbf{114.44} & 10.97 & \textbf{71.94} & 4/7 \\
\addlinespace[2pt]
churro-3B & Page only & 19.32 & 8.73 & 3.13 & 2.10 & 27.26 & 6.91 & 18.00 & \\
 & Multitask (ours) & 22.17 & \textbf{5.48} & \textbf{1.65} & \textbf{2.00} & \textbf{26.81} & \textbf{5.47} & \textbf{17.15} & 6/7 \\
\addlinespace[2pt]
PaddleOCR-VL & Page only & 64.00 & 25.01 & 23.90 & 5.95 & 164.10 & 42.03 & 94.61 & \\
 & Multitask (ours) & \textbf{50.98} & \textbf{21.51} & \textbf{1.05} & \textbf{4.86} & \textbf{63.72} & \textbf{10.18} & \textbf{38.52} & 7/7 \\
\addlinespace[2pt]
DeepSeek-OCR & Page only & 63.45 & 59.41 & 13.06 & 8.09 & 1028.81 & 28.92 & 393.28 & \\
 & Multitask (ours) & 175.16 & \textbf{51.40} & 16.42 & 15.56 & \textbf{393.74} & \textbf{20.60} & 397.73 & 3/7 \\
\addlinespace[2pt]
\midrule
\multicolumn{10}{@{}l}{\textit{Multi-source fine-tuning}} \\
Qwen3.5-2B & Page only & 34.05 & 14.26 & 6.40 & 1.78 & 45.84 & 5.94 & 40.93 & \\
 & Multitask (ours) & \textbf{30.99} & \textbf{8.99} & \textbf{3.24} & 3.18 & 48.07 & \textbf{3.42} & \textbf{39.94} & 5/7 \\
\addlinespace[2pt]
churro-3B & Page only & 44.53 & 5.67 & 4.34 & 1.66 & 11.98 & 5.17 & 22.96 & \\
 & Multitask (ours) & \textbf{16.83} & \textbf{3.18} & \textbf{1.93} & \textbf{1.59} & \textbf{10.81} & \textbf{3.25} & \textbf{18.94} & 7/7 \\
\addlinespace[2pt]
PaddleOCR-VL & Page only & 49.12 & 8.66 & 4.55 & 2.60 & 34.55 & 8.43 & 48.01 & \\
 & Multitask (ours) & 49.37 & 14.50 & 5.59 & 2.67 & \textbf{33.34} & 8.71 & \textbf{43.34} & 2/7 \\
\addlinespace[2pt]
\bottomrule
\end{tabular*}
\end{table*}

With one model per dataset, training on our data lowers CER in 20 of the 28
model--dataset pairs. The benefit is most consistent for PaddleOCR-VL, which
improves on all seven collections, most strongly on HWDB2.0 (23.90\% to
1.05\%), RIMES (42.03\% to 10.18\%), and READ-2016 (164.10\% to 63.72\%), and
for churro-3B, which improves on six of seven. The largest reductions occur
where training on the original annotations fails to produce stable output:
on READ-2016, CER falls from 317.80\% to 114.44\% for Qwen3.5-2B and from
1028.81\% to 393.74\% for DeepSeek-OCR, consistent with the role of the
line-level tasks in stabilizing long generation. The data does not help
uniformly, however: DeepSeek-OCR degrades on four collections, including
BRESSAY (63.45\% to 175.16\%).

With one model for all datasets, the multitask data improves 14 of 21
pairs, but less uniformly. churro-3B improves on all seven collections, most
strongly on BRESSAY (44.53\% to 16.83\%) and HWDB2.0 (4.34\% to 1.93\%), and
Qwen3.5-2B improves on five, including Bentham (14.26\% to 8.99\%), HWDB2.0
(6.40\% to 3.24\%), and RIMES (5.94\% to 3.42\%), but degrades on IAM and
READ-2016. PaddleOCR-VL improves only on READ-2016 and ScribbleLens and
degrades slightly elsewhere. Overall, the multitask data lowers CER in 34 of
49 model--dataset pairs. It helps in both settings, but its effect depends on
the model: it is consistent for churro-3B and for single-source PaddleOCR-VL,
while DeepSeek-OCR and multi-source PaddleOCR-VL benefit only on some
collections. Because the two versions differ in the full set of targets, this
comparison measures the effect of the complete multitask data rather than of
individual tasks, which are ablated in Sec.~\ref{sec:ablation}.

\subsection{Transfer to Other Backbones and Integration Cost}
\label{sec:backbones}

To test whether PF-MoE depends on the Qwen3.5-0.8B backbone, we insert it
into two larger VLMs: churro-3B, a Qwen2.5-VL model adapted to historical
documents \citep{churro_htr}, and Qwen3.5-2B. For each, the baseline is the LoRA checkpoint with the lowest validation CER on our
multitask data, with the LoRA weights merged. PF-MoE is inserted into this checkpoint (churro-3B: layers
11, 17, 23, and 29; Qwen3.5-2B: layers 7, 11, 15, and 19) and fine-tuned with
LoRA on the language model for one epoch, while the backbone weights stay
frozen.

\begin{table}[tbp]
\centering
\caption{
PF-MoE on three backbones (test split, 1,327 pages). \textit{Train.}:
parameters trained in the last stage; \textit{Inf.}: parameters at inference
(LoRA merged, frozen ConvNeXt included), both in millions. Stuck: pages whose
generation reaches the token limit. On churro-3B and
Qwen3.5-2B, PF-MoE is fine-tuned from the baseline checkpoint.
}
\label{tab:backbones}

\footnotesize
\setlength{\tabcolsep}{2.5pt}
\renewcommand{\arraystretch}{1.05}

\begin{tabular*}{\linewidth}
{@{\extracolsep{\fill}} l rr rrr r @{}}
\toprule
Model & Train. & Inf. & CER & WER & Macro CER & Stuck \\
\midrule
\multicolumn{7}{@{}l}{\textit{Qwen3.5-0.8B (full fine-tuning)}} \\
Dense & 852.7 & 852.7 & 8.81 & 19.29 & 11.57 & 16 \\
+ PF-MoE & 933.6 & 961.4 & \textbf{6.81} & \textbf{15.98} & \textbf{8.67} & \textbf{3} \\
\midrule
\multicolumn{7}{@{}l}{\textit{churro-3B (LoRA)}} \\
LoRA & 148.6 & 3755 & 10.55 & 22.19 & 8.08 & 19 \\
+ PF-MoE & 583.8 & 4247 & \textbf{8.51} & \textbf{17.82} & \textbf{6.24} & \textbf{7} \\
\midrule
\multicolumn{7}{@{}l}{\textit{Qwen3.5-2B (LoRA)}} \\
LoRA & 43.6 & 2213 & 19.95 & \textbf{32.52} & 19.69 & 62 \\
+ PF-MoE & 308.5 & 2506 & \textbf{18.19} & 32.53 & \textbf{15.86} & \textbf{46} \\
\bottomrule
\end{tabular*}
\end{table}

Table~\ref{tab:backbones} summarizes the results. On churro-3B, PF-MoE
reduces CER from 10.55\% to 8.51\%, WER from 22.19\% to 17.82\%, and the
number of pages that reach the token limit from 19 to 7. It improves
five of the seven collections, most strongly ScribbleLens (18.94\% to
10.80\%), BRESSAY (16.83\% to 13.14\%), and HWDB2.0 (1.93\% to 1.55\%), leaves
IAM unchanged, and degrades RIMES (3.25\% to 4.48\%). On Qwen3.5-2B, CER
decreases from 19.95\% to 18.19\% and macro-averaged CER from 19.69\% to
15.86\%, and the number of pages that reach the token limit falls from
62 to 46, while WER is unchanged. Most of the remaining failures, 42 of the
46 pages, come from BRESSAY.
On both backbones, the changes resemble those on Qwen3.5-0.8B: CER and macro-averaged CER decrease,
and fewer pages reach the token limit, so the effect of PF-MoE does not
depend on a particular backbone family, model size, or fine-tuning method.

\paragraph{Integration cost.} PF-MoE requires only two properties of the host
VLM: decoder layers with explicit query--key attention over the visual tokens,
from which the pointer prior is read, and a known pixel footprint of each
visual token, to which the fine-grained cells are aligned.
Table~\ref{tab:integration} lists how it is attached to the three backbones.
On the hybrid Qwen3.5 decoders, it occupies four of the six full-attention
layers; on churro-3B, whose Qwen2.5-VL decoder uses full attention in every
layer, it is placed in four evenly spaced layers. The pointer adapts to the
attention variant of each model (query and key normalization in Qwen3.5,
multimodal rotary encoding in Qwen2.5-VL), and for churro-3B, whose visual
tokens cover $28\times28$ pixels, the page is resampled for the ConvNeXt so
that the cells stay aligned with the tokens. The expert width is set so that
the four experts together match the width of the original FFN. No other part
of the backbone is changed, and the same module works with full fine-tuning
and with LoRA. The added module is small relative to the backbone, about 13\%
of its parameters on all three models.

\begin{table}[tbp]
\centering
\caption{
Attaching PF-MoE to three backbones. \textit{Added}: parameters added by
PF-MoE in millions (including the 27.8M frozen ConvNeXt parameters).
}
\label{tab:integration}

\footnotesize
\setlength{\tabcolsep}{2.5pt}
\renewcommand{\arraystretch}{1.05}

\begin{tabular*}{\linewidth}
{@{\extracolsep{\fill}} l l l r r @{}}
\toprule
Backbone & PF-MoE layers & Training & Expert width & Added \\
\midrule
Qwen3.5-0.8B & 7, 11, 15, 19 & full & $4\times896$ & 108.6 \\
Qwen3.5-2B & 7, 11, 15, 19 & LoRA & $4\times1536$ & 292.7 \\
churro-3B & 11, 17, 23, 29 & LoRA & $4\times2752$ & 491.9 \\
\bottomrule
\end{tabular*}
\end{table}

\subsection{Ablation Studies}
\label{sec:ablation}

We next examine the main design choices of ExpertHTR. We first study the
effect of complementary multitask supervision within each dataset and then
compare the resulting single-source models with the jointly trained Dense
model. We use RIMES and HWDB2.0 to examine the roles of the individual
auxiliary tasks and finally ablate the components of PF-MoE on all seven
datasets.

\paragraph{Multitask supervision and multi-source training.}
Table~\ref{tab:mtl_multisource} compares Page-Transcription-only training,
single-source multitask learning, and multi-source multitask learning. The
first two settings are trained independently on each dataset, whereas the
multi-source model, Dense, is trained jointly on all seven sources using the
same multitask formulation. For the comparison between Page-Transcription-only
and single-source multitask training, both variants use the same
optimizer-update budget. The multitask models therefore do not receive
additional parameter updates simply because they contain more task records.

\begin{table}[tbp]
\centering
\caption{
Effect of complementary multitask supervision and multi-source training.
All values are CER (\%). Page-Transcription-only and single-source multitask
models are matched in optimizer-update budget. The multi-source model is
Dense, trained for four epochs in total (Sec.~\ref{sec:experimental_setup}). The best and second-best results for each
dataset are shown in bold and underlined, respectively.
}
\label{tab:mtl_multisource}

\footnotesize
\setlength{\tabcolsep}{2.5pt}
\renewcommand{\arraystretch}{1.05}

\begin{tabular*}{\linewidth}
{@{\extracolsep{\fill}} lrrrrrrr @{}}
\toprule
Setting &
BRES. &
Bent. &
HWDB &
IAM &
READ &
RIMES &
Scrib. \\
\midrule

Page Transcription only
& 33.62
& 37.71
& 4.22
& 3.17
& 190.46
& 7.49
& 90.60 \\

Single-source MTL
& \underline{28.23}
& \underline{26.91}
& \underline{2.94}
& \textbf{2.99}
& \underline{134.06}
& \underline{6.31}
& \underline{79.80} \\

Multi-source MTL (Dense)
& \textbf{11.16}
& \textbf{15.40}
& \textbf{2.36}
& \textbf{2.99}
& \textbf{22.67}
& \textbf{4.02}
& \textbf{22.38} \\

\bottomrule
\end{tabular*}
\end{table}

Under the same optimization budget, multitask supervision improves
Page-Transcription-only training on all seven datasets. The largest reductions
occur on READ-2016, Bentham, and BRESSAY, where CER decreases from 190.46\%
to 134.06\%, from 37.71\% to 26.91\%, and from 33.62\% to 28.23\%,
respectively. Improvements are also observed on IAM, HWDB2.0, RIMES, and
ScribbleLens. Because the two settings use the same number of optimizer
updates, the improvement cannot be explained simply by a longer optimization
schedule.

The multi-source model performs better than the corresponding
single-source multitask models on six of the seven datasets and equally well
on IAM (2.99\%). The largest differences occur on the historical collections:
CER decreases from 134.06\% to 22.67\% on READ-2016, from 79.80\% to 22.38\%
on ScribbleLens, from 28.23\% to 11.16\% on BRESSAY, and from 26.91\% to
15.40\% on Bentham. RIMES (6.31\% to 4.02\%) and HWDB2.0 (2.94\% to 2.36\%)
also improve, so joint training with six Latin-script collections does not
harm the only Chinese collection.

Training one model on all collections is therefore more effective than
training the same multitask formulation separately on each of them. Since
the multi-source model sees the union of the seven training sets, this
comparison reflects the practical benefit of the single-checkpoint setting
adopted by ExpertHTR; it does not isolate transfer between particular
sources. This multi-source model serves as the Dense baseline in the
PF-MoE experiments below.

\paragraph{Auxiliary-task ablation.}
We next examine the roles of the three auxiliary tasks on two
single-source settings with different scripts and annotation structures:
RIMES (French letters, multiple regions per page) and HWDB2.0 (Chinese,
one region per page). Page Transcription is retained in every configuration,
while Physical Line Counting (PLC), Text-to-Line Localization (T2L), and
Line-Span Transcription (LST) are removed individually.

\begin{table}[tbp]
\centering
\caption{
Auxiliary-task ablation on RIMES and HWDB2.0 (single-source training, test
sets of 273 and 415 pages). PLC, T2L, and LST denote Physical Line Counting,
Text-to-Line Localization, and Line-Span Transcription. WER is not reported
for HWDB2.0 because no Chinese word segmentation is applied. The best and
second-best results in each column are shown in bold and underlined.
}
\label{tab:task_ablation}

\footnotesize
\setlength{\tabcolsep}{4pt}
\renewcommand{\arraystretch}{1.05}

\begin{tabular*}{\linewidth}
{@{\extracolsep{\fill}} lccc rr r @{}}
\toprule
& & & & \multicolumn{2}{c}{RIMES} & HWDB2.0 \\
\cmidrule(lr){5-6}\cmidrule(l){7-7}
Setting & PLC & T2L & LST & CER $\downarrow$ & WER $\downarrow$ & CER $\downarrow$ \\
\midrule
Page Transcription only & -- & -- & -- & 7.49 & 18.32 & 4.22 \\
w/o LST & $\checkmark$ & $\checkmark$ & -- & \underline{6.49} & \underline{16.34} & 3.62 \\
w/o T2L & $\checkmark$ & -- & $\checkmark$ & 7.10 & 16.37 & 3.39 \\
w/o PLC & -- & $\checkmark$ & $\checkmark$ & 8.19 & 17.59 & \textbf{2.85} \\
Full multitask & $\checkmark$ & $\checkmark$ & $\checkmark$ & \textbf{6.31} & \textbf{15.77} & \underline{2.94} \\
\bottomrule
\end{tabular*}
\end{table}

Table~\ref{tab:task_ablation} shows that the complete multitask formulation
improves substantially over Page Transcription alone on both datasets. On
RIMES, CER decreases from 7.49\% to 6.31\% and WER from 18.32\% to 15.77\%;
on HWDB2.0, CER decreases from 4.22\% to 2.94\%, a relative reduction of
30\%. On HWDB2.0, every configuration with auxiliary tasks outperforms Page
Transcription alone.

The contribution of individual tasks, however, depends on the dataset.
Removing PLC causes the largest degradation on RIMES, where CER rises to
8.19\%, above Page Transcription alone, whereas on HWDB2.0 the same
configuration gives the lowest CER (2.85\%), 0.09 points below the full
formulation. Conversely, removing LST changes RIMES least but causes the
largest degradation on HWDB2.0 (3.62\%). The effects are therefore not
additive, and the tasks interact during joint training rather than
contributing independent gains that can be ranked in isolation. The full
formulation is the best configuration on RIMES and within 0.09 points of the
best on HWDB2.0, and it is the only configuration among the best two in every
column. We therefore use all four tasks in the remaining experiments.

\paragraph{Components of PF-MoE.}
We finally ablate the components of PF-MoE on all seven datasets. Each
variant removes or replaces one component of the full model and is trained
with the same data and number of optimizer updates as ExpertHTR. To make the
comparison self-explanatory, we name each variant by what it can still do
rather than by the component that is removed:
\begin{itemize}
    \item \emph{Dense}: the dense multitask model
    (Sec.~\ref{sec:experimental_setup}), trained from the same three-epoch
    checkpoint for one further epoch without PF-MoE; it has no experts.
    \item \emph{Text-only experts}: four routed experts in the same layers,
    without the fine-grained evidence pathway; the experts see only
    $\mathbf{x}_t$.
    \item \emph{Single expert}: the complete evidence pathway, but one expert
    of width 3584 instead of four routed experts of width 896, with a
    similar number of parameters.
    \item \emph{Random cells}: the pointer is replaced by 16 cells chosen
    at random, with uniform weights.
    \item \emph{Unrefined cells}: the local refinement of the cell encoder
    is removed, so the cells are linear projections of the ConvNeXt
    features.
    \item \emph{Evidence off at test}: the ExpertHTR checkpoint evaluated
    with the evidence disabled ($\mathbf{u}_t=\mathbf{0}$) at inference.
\end{itemize}

\begin{table*}[!t]
\centering
\caption{
Component ablation of PF-MoE on the seven test sets (Qwen3.5-0.8B). The columns \textit{Experts}--\textit{Refine.} indicate which
parts of the model are active. \textit{Added} is the number of parameters
added to the 852.74M of the backbone, including the 27.8M parameters of the
frozen ConvNeXt stages (80.8M of the 108.6M are trained in the full model).
CER and WER are micro-averaged over all 1,327 test pages (WER without
HWDB2.0). Macro CER is the unweighted mean of the
seven per-dataset CERs. Skip.\ and Rep.: skipped and repeated lines per 100
reference lines; Stuck: pages whose generation reaches the token limit. The
best and second-best results in each column are shown in bold and underlined.
}
\label{tab:moe_ablation}

\scriptsize
\setlength{\tabcolsep}{3pt}
\renewcommand{\arraystretch}{1.08}

\begin{tabular*}{\textwidth}
{@{\extracolsep{\fill}} l cccc r r r r rrr @{}}
\toprule
Variant & Experts & Cells & Pointer & Refine. & Added (M) & CER $\downarrow$ & WER $\downarrow$ & Macro CER $\downarrow$ & Skip. $\downarrow$ & Rep. $\downarrow$ & Stuck $\downarrow$ \\
\midrule
Dense & -- & -- & -- & -- & 0.0 & 8.81 & 19.29 & 11.57 & 1.79 & 3.31 & 16 \\
Text-only experts & 4 & -- & -- & -- & 80.4 & 7.39 & 17.59 & 9.61 & 1.52 & 0.23 & 7 \\
Single expert & 1 & $\checkmark$ & $\checkmark$ & $\checkmark$ & 105.5 & 7.25 & 17.04 & \underline{8.27} & 1.61 & \underline{0.12} & 9 \\
Random cells & 4 & $\checkmark$ & random & $\checkmark$ & 108.0 & \underline{7.05} & \underline{16.71} & \textbf{8.22} & \underline{1.21} & 0.94 & \underline{5} \\
Unrefined cells & 4 & $\checkmark$ & $\checkmark$ & -- & 108.2 & 7.78 & 17.99 & 10.13 & 1.72 & 3.26 & 12 \\
Evidence off at test & 4 & off & off & off & 108.6 & 7.46 & 17.00 & 8.79 & 1.33 & 0.24 & 7 \\
\midrule
ExpertHTR (full) & 4 & $\checkmark$ & $\checkmark$ & $\checkmark$ & 108.6 & \textbf{6.81} & \textbf{15.98} & 8.67 & \textbf{1.01} & \textbf{0.11} & \textbf{3} \\
\bottomrule
\end{tabular*}
\end{table*}

\begin{table*}[!t]
\centering
\caption{
Per-dataset CER and WER (\%) of the component ablation. WER is not reported
for HWDB2.0. The four largest collections contain 1,223 of the 1,327 test
pages; Bentham, READ-2016, and ScribbleLens contain 33, 50, and 21. The best and
second-best results in each column are shown in bold and underlined.
}
\label{tab:moe_ablation_datasets}

\scriptsize
\setlength{\tabcolsep}{2pt}
\renewcommand{\arraystretch}{1.08}

\begin{tabular*}{\textwidth}
{@{\extracolsep{\fill}} l rr rr r rr rr rr rr @{}}
\toprule
& \multicolumn{2}{c}{BRESSAY} & \multicolumn{2}{c}{Bentham} & HWDB2.0
& \multicolumn{2}{c}{IAM} & \multicolumn{2}{c}{READ-2016}
& \multicolumn{2}{c}{RIMES} & \multicolumn{2}{c}{ScribbleLens} \\
\cmidrule(lr){2-3}\cmidrule(lr){4-5}\cmidrule(lr){6-6}\cmidrule(lr){7-8}
\cmidrule(lr){9-10}\cmidrule(lr){11-12}\cmidrule(l){13-14}
Variant & CER & WER & CER & WER & CER & CER & WER & CER & WER & CER & WER & CER & WER \\
\midrule
Dense & 11.16 & 20.55 & 15.40 & 28.02 & 2.36 & 2.99 & \textbf{7.43} & 22.67 & 53.98 & \underline{4.02} & \underline{10.92} & 22.38 & 51.58 \\
Text-only experts & 9.02 & 17.86 & 15.39 & 30.79 & 2.29 & 2.99 & 7.69 & 17.82 & 50.90 & 4.03 & 11.02 & 15.73 & 41.44 \\
Single expert & 9.49 & 18.75 & \textbf{4.95} & \textbf{10.88} & 2.33 & 3.21 & 8.11 & 19.48 & 46.56 & \textbf{3.96} & \textbf{10.90} & \textbf{14.44} & \textbf{40.50} \\
Random cells & \underline{8.89} & \underline{17.76} & \underline{8.99} & 18.49 & \underline{2.26} & 3.05 & 7.83 & 14.50 & \textbf{39.39} & 4.33 & 11.23 & 15.53 & \underline{41.42} \\
Unrefined cells & 9.40 & 18.95 & 21.12 & 29.56 & \textbf{2.22} & 3.11 & 7.72 & \underline{13.98} & \underline{39.59} & 4.18 & 11.37 & 16.89 & 44.32 \\
Evidence off at test & 9.45 & 18.59 & 13.51 & \underline{14.79} & 2.42 & \underline{2.97} & \underline{7.66} & \textbf{13.79} & 39.93 & 4.08 & 10.98 & 15.29 & 42.63 \\
\midrule
ExpertHTR (full) & \textbf{8.17} & \textbf{16.75} & 13.49 & 14.89 & 2.50 & \textbf{2.96} & \underline{7.66} & 14.35 & 40.38 & 4.05 & 10.95 & \underline{15.18} & 42.07 \\
\bottomrule
\end{tabular*}
\end{table*}

Table~\ref{tab:moe_ablation} summarizes the results. PF-MoE as a whole
improves the dense model: CER falls from 8.81\% to 6.81\% and WER from
19.29\% to 15.98\%. Every component contributes to this gain, since removing
any of them raises CER, by 0.24 points without the learned pointer (Random
cells), 0.44 points without routing (Single expert), and 0.58 points without
the evidence (Text-only experts). The local refinement of the cells is
necessary: without it (Unrefined cells), CER rises to 7.78\%, above the
7.39\% of Text-only experts, so unrefined cells are worse than no cells at
all. The trained model also relies on the evidence, since disabling it at
test time raises CER to 7.46\%.

The components also make long pages more reliable to generate. Without
routing (Single expert), skipped lines increase from 1.01 to 1.61 per 100
reference lines and stuck pages from 3 to 9; with random cells instead of
the pointer, repeated lines increase from 0.11 to 0.94 per 100 reference
lines and stuck pages to 5; and without evidence (Text-only experts), 7
pages get stuck. The full model has the best value of every measure that
aggregates over all 1,327 pages: micro-averaged CER and WER, skipped and
repeated lines, and stuck pages.

Table~\ref{tab:moe_ablation_datasets} reports the per-dataset results. On
the four largest collections, which contain 1,223 of the 1,327 test pages,
the full model has the lowest CER on BRESSAY and IAM and is within 0.3
points of the best variant on HWDB2.0 and RIMES. On Bentham, READ-2016, and
ScribbleLens, which together contain 104 pages, a single page with a long
repeated or omitted passage changes the CER of a collection by several
points, and the ranking of the variants varies accordingly; the lower
macro-averaged CER of Single expert and Random cells (8.27\% and 8.22\%
versus 8.67\%) stems almost entirely from Bentham, whose 33 pages make up
less than 3\% of the test set.

Because all variants are compared on the test set, we do not choose among
them by the columns they win in Table~\ref{tab:moe_ablation_datasets}, which
would amount to selecting a model on the test data. We keep the full model
for three reasons: it is the best variant on every measure aggregated over
all test pages; it costs about as many parameters as the alternatives
(105.5M to 108.6M); and its learned pointer gives an explicit image location
for every generated token, which random cells cannot provide. We use this
location, for example, to cut the line images in
Figs.~\ref{fig:cases_fixed} and~\ref{fig:cases_failed} without any line
annotation.

\subsection{Analysis of PF-MoE}
\label{sec:analysis}

The ablations show that the evidence pathway improves recognition, but not
how. We first examine which cells the pointer selects, how the experts are
used, and which errors change, and then examine individual predictions. All
analyses use the Qwen3.5-0.8B models. Pointer and router statistics are
measured with teacher forcing on 120 test pages spread over the seven
datasets, and error counts on all 1,327 test pages.

\begin{table}[tbp]
\centering
\caption{
Character-level errors per 100 reference characters on all 1,327 test pages
(Qwen3.5-0.8B): substitutions (Sub.), deletions (Del.), and insertions
(Ins.). Line-level failures are reported in Table~\ref{tab:moe_ablation}.
}
\label{tab:analysis_stats}

\footnotesize
\setlength{\tabcolsep}{4pt}
\renewcommand{\arraystretch}{1.05}

\begin{tabular*}{\linewidth}
{@{\extracolsep{\fill}} l rrr @{}}
\toprule
Model & Sub. $\downarrow$ & Del. $\downarrow$ & Ins. $\downarrow$ \\
\midrule
Dense & 3.85 & 2.42 & 2.54 \\
Text-only experts & 3.66 & 1.95 & \textbf{1.78} \\
Unrefined cells & 3.84 & 1.93 & 2.01 \\
ExpertHTR & \textbf{3.38} & \textbf{1.63} & 1.79 \\
\bottomrule
\end{tabular*}
\end{table}

\begin{table}[tbp]
\centering
\caption{
Share of pointer weight (\%) on the cells of each stride, averaged over the
four PF-MoE layers (ExpertHTR, Qwen3.5-0.8B; teacher forcing on 120 test
pages spread over the seven datasets). Datasets are sorted by the share of
the 8-pixel cells.
}
\label{tab:stride_weight}

\footnotesize
\setlength{\tabcolsep}{4pt}
\renewcommand{\arraystretch}{1.05}

\begin{tabular*}{\linewidth}
{@{\extracolsep{\fill}} l rrr @{}}
\toprule
Dataset & 8 px & 16 px & 32 px \\
\midrule
READ-2016    & 89.1 & 9.7  & 1.2 \\
ScribbleLens & 81.4 & 13.5 & 5.0 \\
BRESSAY      & 69.7 & 24.2 & 6.2 \\
HWDB2.0      & 69.7 & 26.2 & 4.2 \\
IAM          & 61.3 & 32.5 & 6.2 \\
Bentham      & 61.1 & 30.2 & 8.7 \\
RIMES        & 60.2 & 32.6 & 7.2 \\
\bottomrule
\end{tabular*}
\end{table}

\begin{table*}[tbp]
\centering
\caption{
Share of tokens (\%) routed to each expert in layer 19 (ExpertHTR,
Qwen3.5-0.8B; teacher forcing on the same 120 test pages as
Table~\ref{tab:stride_weight}, 34,946 tokens). Each column sums to 100;
the largest share in each column is shown in bold. The last four columns
group tokens by type across all datasets. \textit{Gate}: mean value of the
correction gate.
}
\label{tab:expert_share}

\scriptsize
\setlength{\tabcolsep}{3pt}
\renewcommand{\arraystretch}{1.05}

\begin{tabular*}{\textwidth}
{@{\extracolsep{\fill}} l rrrrrrr rrrr @{}}
\toprule
Expert & IAM & RIMES & HWDB & BRES. & Bent. & READ & Scrib. & Digits & \texttt{<del>}/\texttt{<gap>} & Chinese & All \\
\midrule
E1 & 9.6 & 11.1 & 1.6 & 3.9 & 13.1 & 0.3 & 0.4 & 14.0 & 0.0 & 0.8 & 4.8 \\
E2 & 17.7 & 27.9 & \textbf{51.2} & 17.8 & 16.2 & 3.2 & 4.7 & \textbf{47.9} & 1.4 & \textbf{52.9} & 16.9 \\
E3 & \textbf{42.8} & 22.7 & 39.1 & \textbf{49.6} & 24.4 & 1.1 & 2.2 & 12.2 & \textbf{97.2} & 37.4 & 27.0 \\
E4 & 29.9 & \textbf{38.3} & 8.1 & 28.8 & \textbf{46.3} & \textbf{95.4} & \textbf{92.7} & 25.9 & 1.4 & 8.9 & \textbf{51.4} \\
Gate & -- & -- & -- & -- & -- & -- & -- & 0.82 & 0.97 & 0.95 & 0.91 \\
\bottomrule
\end{tabular*}
\end{table*}

\paragraph{The pointer adapts the resolution to the writing.}
The pointer is never told which stride to use, yet its choice is consistent
and depends on the collection (Table~\ref{tab:stride_weight}). The 32-pixel
cells receive less than 9\% of the weight everywhere. They cover the same
region as a visual token, so they add little that the decoder does not
already see, and the pointer learns to spend its 16 cells on finer views
instead. The 8-pixel cells dominate in every collection, and their share is
highest where the distinguishing detail is smallest: in the small, faded
historical scripts of READ-2016 (89\%) and ScribbleLens (81\%), and in the
dense Chinese characters of HWDB2.0 and the essays of BRESSAY (70\%). On
the modern Latin collections IAM and RIMES and on Bentham, about a third of
the weight moves to 16-pixel cells, which cover whole letters rather than
stroke fragments. The two collections with the strongest preference for
8-pixel cells are also those on which PF-MoE reduces CER most
(Table~\ref{tab:moe_ablation_datasets}: 22.67\% to 14.35\% on READ-2016 and
22.38\% to 15.18\% on ScribbleLens). The share of fine cells alone does not
predict the gain, however: HWDB2.0 relies on them as much as BRESSAY but
does not improve, which agrees with the finding below that the evidence
mainly stabilizes generation rather than sharpening the recognition of
individual characters.

\paragraph{Experts specialize without supervision.}
PF-MoE uses routed experts because the training mixture is heterogeneous
(Sec.~\ref{sec:introduction}). The router receives no language, script,
dataset, or task identity, yet it recovers this heterogeneity on its own.
Table~\ref{tab:expert_share} shows this for layer 19, the last PF-MoE
layer. Almost all tokens of the two difficult historical collections,
READ-2016 (95\%) and ScribbleLens (93\%), go to E4; HWDB2.0 and Chinese
characters go mainly to E2 (about 52\%); and the \texttt{<del>} and
\texttt{<gap>} markers go almost exclusively to E3 (97\%). The load-balancing loss is a soft
constraint, so the experts are not used equally: E1 receives only 4.8\% of
the tokens, but it still handles part of the digits (14\%) and of the
Bentham and RIMES tokens (13\% and 11\%). The experts
therefore divide the work both by script and by token type, and one of them
is reserved almost entirely for cancelled and illegible content, the tokens
for which removing the evidence matters most (see below). The correction
gate is open for most tokens (mean 0.91) and is highest for these markers
(0.97) and for Chinese characters (0.95), tokens that are hard to predict
from context and for which the decoder has to rely most on what is written.

\paragraph{Which errors change.} The experts and the evidence reduce different
errors. Relative to Dense, ExpertHTR has fewer substitutions (3.85 to 3.38),
deletions (2.42 to 1.63), and insertions (2.54 to 1.79) per 100 reference
characters (Table~\ref{tab:analysis_stats}), fewer skipped (1.79 to 1.01) and
repeated lines (3.31 to 0.11) per 100 reference lines, and fewer pages that
reach the token limit (16 to 3; Table~\ref{tab:moe_ablation}). The added
experts alone, without any evidence (Text-only experts), already remove most
repeated lines and excess insertions, so the extra capacity mainly prevents
the model from running on. The evidence pathway adds the further reduction of
substitutions, deletions, and skipped lines, the errors in which written
content is misread or lost. Both effects depend on the local refinement:
without it, repeated lines and stuck pages return almost to the level of
Dense.

\begin{figure*}[tbp]
    \centering
    \includegraphics[width=\textwidth]{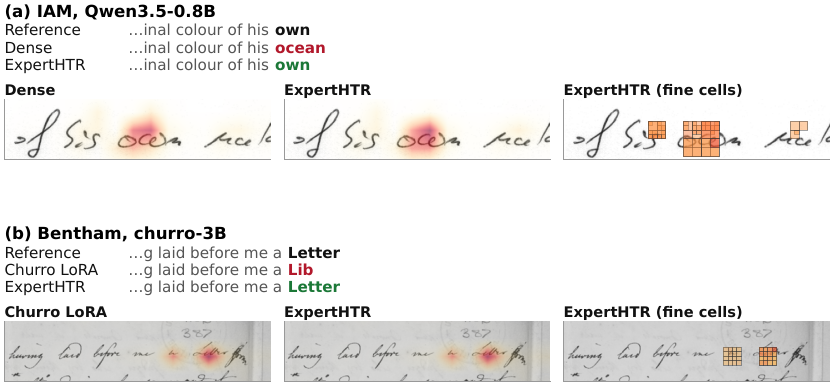}
    \caption{
        Where the models look when the baseline errs (teacher forcing on the
        reference; one generated token each). (a) IAM, Qwen3.5-0.8B: Dense
        predicts \textit{ocean} for \textit{own}. (b) Bentham, churro-3B: the
        LoRA baseline predicts \textit{Lib\ldots} for \textit{Letter}. Left and
        middle: attention of the decoder from the generating position to the
        visual tokens of the baseline and of ExpertHTR (mean over heads and the
        four PF-MoE layers). Right: the fine-grained cells selected by the
        pointer of ExpertHTR (8/16/32 px; color: pointer weight).
    }
    \label{fig:attention}
\end{figure*}

\paragraph{Where the models look.}
Figure~\ref{fig:attention} examines two tokens that the baseline predicts
incorrectly and ExpertHTR correctly, on two backbones. The coarse attention
of the baseline and of ExpertHTR is nearly identical and is centered on the
word to be read, so the baseline does not fail because it looks at the wrong
place. At that place, however, a visual token covers 32 pixels (28 on
churro-3B) and merges several strokes, such as the \textit{w} of
\textit{own} or the \textit{Le} of \textit{Letter}, and the decoder completes
the word from its language prior (\textit{ocean}, \textit{Lib\ldots}). The
pointer of ExpertHTR selects 8- and 16-pixel cells on exactly these strokes.
Under teacher forcing, the probability of the correct token rises from 0.05
to 0.94 for \textit{own}, but only from 0.08 to 0.25 for \textit{Letter}.

\begin{figure*}[tbp]
    \centering
    \includegraphics[width=\textwidth]{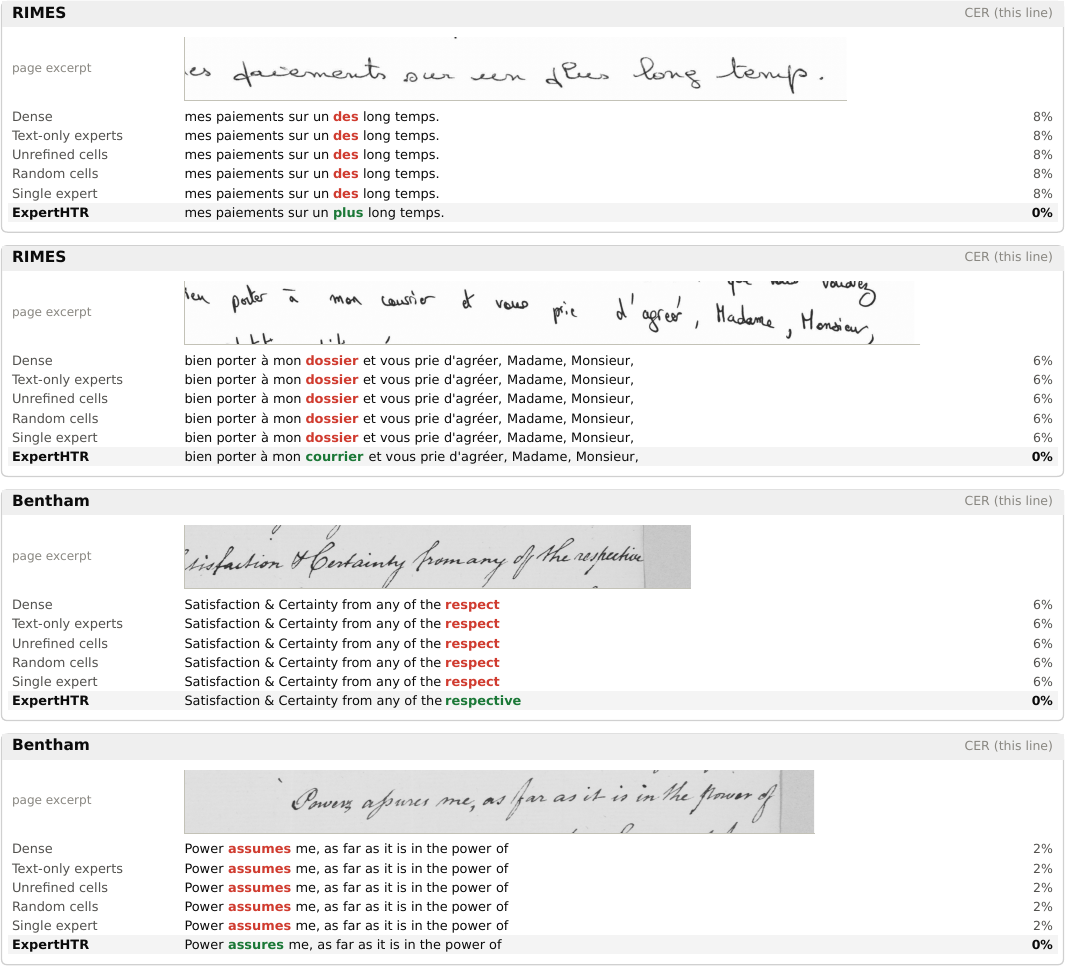}
    \caption{
        Test lines that only ExpertHTR transcribes exactly, with the
        predictions of Dense and of the four ablations. Red: wrong word. The
        percentage on the right is the CER of that line: the number of character
        insertions, deletions, and substitutions needed to turn the predicted
        line into the reference line, divided by the number of characters of the
        reference line. Line crops are cut from the original pages at the
        position given by ExpertHTR's teacher-forced pointer; no line
        annotation is used, and all models read the complete page.
    }
    \label{fig:cases_fixed}
\end{figure*}

\paragraph{Errors that only ExpertHTR avoids.}
Figure~\ref{fig:cases_fixed} shows test lines that only ExpertHTR
transcribes exactly. In each case, Dense and all four ablations produce the
same wrong word, which indicates a shared limitation
rather than noise of individual models. The errors are of two kinds. Some
misread the glyphs, such as \textit{des} for \textit{plus}, which makes the
sentence ungrammatical. Others replace a clearly written word with a fluent
alternative that fits the context, such as \textit{dossier} for
\textit{courrier}, or with a shorter or more frequent form that ignores
final or internal strokes, such as \textit{respect} for \textit{respective}
and \textit{assumes} for \textit{assures}. The ablations show that access to
fine-grained cells alone is not sufficient: Unrefined cells, Random cells,
and Single expert, which all receive fine-grained evidence, make the same
errors, and only the combination of the learned pointer, the local
refinement, and routed experts recovers these words.

\paragraph{How representative these cases are.}
Figures~\ref{fig:attention} and~\ref{fig:cases_fixed} show selected
examples. In aggregate, the evidence matters less for the likelihood of
individual characters than these cases suggest. When we remove the evidence
from the trained ExpertHTR model ($\mathbf{u}_t=\mathbf{0}$) and keep
routing, gate, and pointer unchanged, the mean per-token negative
log-likelihood under teacher forcing increases only for the \texttt{<del>}
and \texttt{<gap>} markers (+0.038 nats, 95\% interval $[+0.026,+0.051]$) and
changes by less than 0.002 nats for letters, digits, punctuation, and
Chinese characters. In free-running decoding, the same removal raises CER
from 6.81\% to 7.46\% and the number of stuck pages from 3 to 7. The cases
above are therefore instances in which the fine-grained evidence is
decisive for a word, whereas most of the aggregate gain comes from more
stable generation, with fewer skipped lines, repetition loops, and
deletions. About 70\% of all word substitutions remain substitutions by
another real word for every model, and the difference between ExpertHTR
(7.72 per 100 words) and Dense (8.00) in such substitutions is not
statistically significant.

\subsection{Error Analysis}
\label{sec:error_analysis}

\begin{figure*}[tbp]
    \centering
    \includegraphics[width=\textwidth]{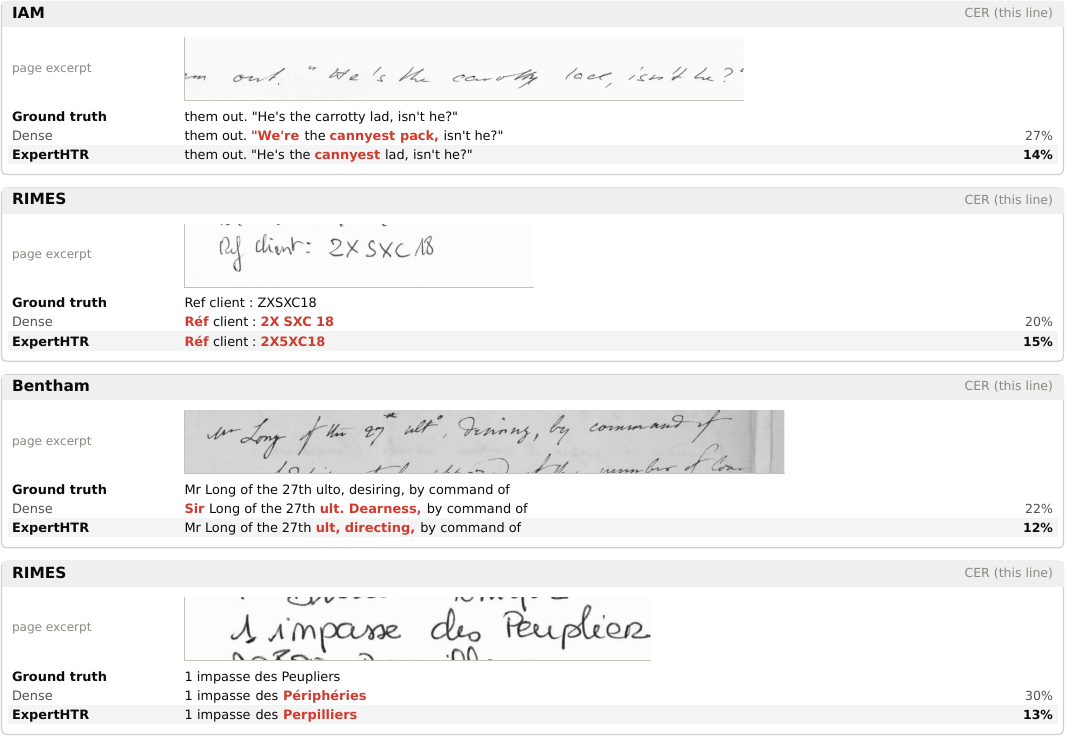}
    \caption{
        Test lines that ExpertHTR still transcribes incorrectly, compared with
        Dense. Red: wrong characters or words. The percentage on the right is
        the CER of that line, defined as in Fig.~\ref{fig:cases_fixed}; the line
        crops are obtained in the same way.
    }
    \label{fig:cases_failed}
\end{figure*}

\paragraph{Remaining errors.}
Figure~\ref{fig:cases_failed} shows lines that ExpertHTR still gets wrong.
The two models fail differently. Dense replaces whole words with common
ones (\textit{P\'eriph\'eries}, \textit{pack}, \textit{Dearness}, \textit{Sir}),
whereas ExpertHTR keeps the shape of the word and errs on a few characters
(\textit{Perpilliers} for \textit{Peupliers}, \textit{cannyest} for
\textit{carrotty}, \textit{directing} for \textit{desiring}), and its line
CER is lower than that of Dense in every case. Most of these errors
originate in the data rather than in the visual pathway: rare words that do not occur
in the training vocabulary (the street name \textit{Peupliers}, the dialect
word \textit{carrotty}), handwriting that is ambiguous in itself (in the
customer code \textit{ZXSXC18}, Z and S are written like 2 and 5), and
historical conventions such as the abbreviation \textit{ulto} with a
superscript \textit{o}. Finer evidence helps to reproduce the written shape
more faithfully but cannot supply missing lexical knowledge.

\paragraph{Line-level outcome.}
Matching each reference line to its closest predicted line shows where the
remaining character errors occur. Per 100 reference characters, ExpertHTR
has 4.4 errors on lines with large errors (Dense: 4.8), 1.8 on lines with
small errors (Dense: 1.7), and 0.6 from skipped lines (Dense: 1.9), and the
share of lines transcribed exactly changes only from 49.8\% to 50.4\%.
ExpertHTR thus does not transcribe more lines perfectly; it loses fewer
lines and fails less severely on the hardest ones.

\section{Discussion}
\label{sec:discussion}

The experiments evaluate two components that act at different levels: the
structure of the training data and the access of the decoder to fine-grained
visual evidence. This section interprets what each contributes, how ExpertHTR
relates to specialized systems, and where the approach is limited.

\subsection{Structured Supervision from Existing Annotations}
\label{sec:discussion_supervision}

The main lesson of the supervision experiments is that existing page
annotations contain more training signal than a single transcription target
uses. Re-expressing the same pages as four tasks requires no new labels, yet
it improves Page-Transcription-only training in every single-source setting
and helps fine-tuned OCR/VLM models of other architectures in most
model--dataset pairs (Secs.~\ref{sec:data_effect}
and~\ref{sec:ablation}). The effect is largest where page-level generation
is least stable, such as READ-2016, which fits the purpose of the
line-level tasks: Physical Line Counting, Text-to-Line Localization, and
Line-Span Transcription turn coverage, position, and line boundaries into
explicit targets instead of leaving them implicit in a long transcription.

The benefit is, however, not uniform. It varies between models in the
multi-source setting, and the auxiliary-task ablation shows that the most
useful task depends on the collection (Table~\ref{tab:task_ablation}). The
multitask data should therefore be seen as an inexpensive and general way to
obtain more supervision from existing annotations, usable both for one model
over many collections and for a model of a single collection, rather than as
a fixed recipe in which every task always helps.

\subsection{What Fine-Grained Evidence Provides}
\label{sec:discussion_evidence}

PF-MoE was motivated by the coarse granularity of VLM visual tokens, and
its measured effect is more specific than this motivation suggested. The
evidence does not, on average, make individual characters more likely; it
helps the model keep its place on long pages and decide whether cancelled or
illegible content should be transcribed (Sec.~\ref{sec:analysis}). This
explains where the gains appear: on the hard-to-read collections, where long
generation fails most often, rather than on clean modern pages, where the
dense model already reads reliably (Sec.~\ref{sec:main_results}). The
examples in Figs.~\ref{fig:attention} and~\ref{fig:cases_fixed} show that
the evidence can still be decisive for individual words, but they are the
exception rather than the rule.

The ablations and the analysis also show that \emph{how} the evidence is
provided matters as much as \emph{whether} it is provided. Unrefined
ConvNeXt features are worse than no evidence at all, so a small amount of
task-specific processing of the cells is a precondition for any benefit.
Without supervision, the pointer adapts the resolution to the writing and,
in the last PF-MoE layer, the experts divide the work by script and token
type, which shows that the
module finds structure in the heterogeneous data on its own. The learned
pointer and routing each lower CER and make long generation more reliable
(Sec.~\ref{sec:ablation}).

\subsection{Positioning with Respect to Specialized Systems}
\label{sec:discussion_positioning}

ExpertHTR and the DAN family represent two different trade-offs. The DAN
family spends its effort on each target collection, with a dedicated model
and synthetic pretraining generated from the text of that collection, and
remains more accurate on the collections where this matters most. ExpertHTR
spends it on generality: one checkpoint covers seven collections, two
scripts, and six languages without synthetic data, and the same module
attaches to VLMs of different families and sizes, with full fine-tuning or
LoRA, at about 13\% additional parameters. That this general model is
already better than or close to the specialized systems on IAM, HWDB2.0,
and RIMES suggests that the two approaches are complementary rather than
competing: target-matched synthetic pretraining could be combined with our
structured supervision when a single collection is of interest.

\subsection{Limitations and Future Directions}
\label{sec:limitations}

Despite these results, four limitations remain. First, the recognizer
inherits the language coverage of its pretrained backbone. Results are
strongest on modern English and French collections (IAM and RIMES), for
which large amounts of text are available during VLM pretraining, and
weakest on historical Dutch and German (ScribbleLens and READ-2016), where
spelling, abbreviations, and vocabulary differ from modern usage. Several
remaining errors are of this kind, such as a rare street name or a dialect
word read as a more common form (Fig.~\ref{fig:cases_failed}).

Second, the fine-grained evidence does not yet overcome this language prior.
Removing it leaves the likelihood of ordinary characters unchanged, and
ExpertHTR substitutes one real word for another about as often as Dense
(Sec.~\ref{sec:analysis}). Once the correct context is given, the evidence
therefore rarely changes which word the decoder chooses. We see two likely
causes. Architecturally, the evidence enters only through a gated, additive
expert update in four of the 24 decoder layers, while the strongly
pretrained remaining layers and the output head continue to favor fluent
continuations. In terms of training, the language-modeling objective can be
minimized largely from context, so nothing forces the model to rely on the
evidence where context and handwriting disagree. How to make a strongly
pretrained decoder follow the written evidence rather than its language
prior, without losing the benefit of that prior where the handwriting is
ambiguous, remains an open problem.

Third, page-level handwriting datasets remain scarce. Our collections
contain between 164 and 1,377 training pages each, and the historical
collections Bentham, READ-2016, and ScribbleLens provide only 21 to 50 test
pages. The multitask formulation extracts more supervision from these
annotations, but it cannot replace additional page-level data, particularly
for historical material; larger historical test sets would also allow finer
comparisons between model variants.

Fourth, languages and scripts beyond those in our training mixture remain a
major barrier. Extending the approach to other scripts, to low-resource
languages, and to documents with complex layouts will require new page-level
annotations and, for many scripts, backbones whose pretraining covers them. Future work includes collecting or
synthesizing page-level data for underrepresented languages, and combining
the structured supervision with target-matched synthetic pretraining or with
backbones pretrained on historical text.

\section{Conclusion}
\label{sec:conclusion}

We presented ExpertHTR, a page-level handwritten text recognition framework
built on pretrained vision--language models. At the supervision level, a
common Page--Region--Line representation turns the existing annotations of
seven heterogeneous collections into four complementary training tasks
without additional labeling. At the model level, the pointer-guided
fine-grained Mixture-of-Experts lets selected decoder layers revisit the
page at 8-, 16-, and 32-pixel resolution at the location being transcribed,
without increasing the number of visual tokens.

A single checkpoint reduces CER from 8.81\% to 6.81\% relative to the dense
multitask baseline, with fewer skipped lines, repeated lines, and repetition
loops, and the module transfers to
churro-3B and Qwen3.5-2B. The structured training data also improves
fine-tuned OCR/VLM models of other architectures, in 34 of 49 model--dataset
pairs across single- and multi-source training. Our analysis shows that the fine-grained
evidence mainly stabilizes long generation and helps identify cancelled and
illegible text, rather than improving the recognition of individual
characters. Specialized systems with dataset-specific synthetic pretraining
remain stronger on several historical collections, and it is still open how
to make the decoder follow the handwriting when it conflicts with its
language prior. These two questions are the main directions for future
work.

%\section*{Declaration of generative AI and AI-assisted technologies in the manuscript preparation process}
%During the preparation of this work the authors used Claude (Anthropic) to
%improve the clarity and readability of the text and to help describe the architecture and data-flow diagrams. After using this tool, the authors
%reviewed and edited the content as needed and take full responsibility for
%the content of the published article.
%
%\section*{Data availability}
%The seven handwriting datasets used in this study are publicly available
%from their original providers. The code and configurations used for the
%experiments are available at \url{https://github.com/DAIR-Group/ExpertHTR}.

%\section*{Acknowledgements}
%This research was supported by The VNUHCM-University of Information
%Technology's Scientific Research Support Fund.

\bibliographystyle{model1-num-names}
\bibliography{cas-refs}

@article{dhiaf2025ucl,
  title={UCL-MHTR: A unified continual learning system for multilingual handwritten text recognition},
  author={Dhiaf, Marwa and Souibgui, Mohamed Ali and Kessentini, Yousri and Fornes, Alicia and Rouhou, Ahmed Cheikh},
  journal={Expert Systems with Applications},
  volume={294},
  pages={128741},
  year={2025},
  publisher={Elsevier}
}

@inproceedings{fujitake2024dtrocr,
  title={Dtrocr: Decoder-only transformer for optical character recognition},
  author={Fujitake, Masato},
  booktitle={2024 IEEE/CVF Winter Conference on Applications of Computer Vision (WACV)},
  pages={8010--8020},
  year={2024},
  organization={IEEE}
}

@article{msdoctr_lite,
  title={MSdocTr-Lite: A lite transformer for full page multi-script handwriting recognition},
  author={Dhiaf, Marwa and Rouhou, Ahmed Cheikh and Kessentini, Yousri and Salem, Sinda Ben},
  journal={Pattern Recognition Letters},
  volume={169},
  pages={28--34},
  year={2023},
  publisher={Elsevier}
}

@article{htr_survey,
  title={Handwritten text recognition: A survey},
  author={Garrido-Munoz, Carlos and Rios-Vila, Antonio and Calvo-Zaragoza, Jorge},
  journal={IEEE Transactions on Pattern Analysis and Machine Intelligence},
  year={2025},
  publisher={IEEE}
}

@inproceedings{span_htr,
  title={SPAN: a simple predict \& align network for handwritten paragraph recognition},
  author={Coquenet, Denis and Chatelain, Cl{\'e}ment and Paquet, Thierry},
  booktitle={International Conference on Document Analysis and Recognition},
  pages={70--84},
  year={2021},
  organization={Springer}
}

@article{dan_htr,
  title={Dan: a segmentation-free document attention network for handwritten document recognition},
  author={Coquenet, Denis and Chatelain, Cl{\'e}ment and Paquet, Thierry},
  journal={IEEE transactions on pattern analysis and machine intelligence},
  volume={45},
  number={7},
  pages={8227--8243},
  year={2023},
  publisher={IEEE}
}

@inproceedings{fasterdan_htr,
  title={Faster dan: Multi-target queries with document positional encoding for end-to-end handwritten document recognition},
  author={Coquenet, Denis and Chatelain, Cl{\'e}ment and Paquet, Thierry},
  booktitle={International Conference on Document Analysis and Recognition},
  pages={182--199},
  year={2023},
  organization={Springer}
}

@article{daniel_htr,
  title={DANIEL: a fast document attention network for information extraction and labelling of handwritten documents: T. Constum et al.},
  author={Constum, Thomas and Tranouez, Pierrick and Paquet, Thierry},
  journal={International Journal on Document Analysis and Recognition (IJDAR)},
  volume={28},
  number={4},
  pages={573--595},
  year={2025},
  publisher={Springer}
}

@article{metadan_htr,
  title={Meta-DAN: towards an efficient prediction strategy for page-level handwritten text recognition},
  author={Coquenet, Denis},
  journal={Pattern Recognition},
  pages={113373},
  year={2026},
  publisher={Elsevier}
}

@inproceedings{htr_generalization,
  title={On the generalization of handwritten text recognition models},
  author={Garrido-Munoz, Carlos and Calvo-Zaragoza, Jorge},
  booktitle={2025 IEEE/CVF Conference on Computer Vision and Pattern Recognition (CVPR)},
  pages={15275--15286},
  year={2025},
  organization={IEEE}
}

@article{crosilla_htr,
  title={Benchmarking large language models for handwritten text recognition},
  author={Crosilla, Giorgia and Klic, Lukas and Colavizza, Giovanni},
  journal={Journal of Documentation},
  volume={81},
  number={7},
  pages={334--354},
  year={2025},
  publisher={Emerald Publishing Limited}
}

@inproceedings{churro_htr,
  title={Churro: Making history readable with an open-weight large vision-language model for high-accuracy, low-cost historical text recognition},
  author={Semnani, Sina and Zhang, Han and He, Xinyan and Tekg{\"u}rler, Merve and Lam, Monica},
  booktitle={Proceedings of the 2025 Conference on Empirical Methods in Natural Language Processing},
  pages={34765--34812},
  year={2025}
}

@article{unimummer,
  title={Uni-mumer: Unified multi-task fine-tuning of vision-language model for handwritten mathematical expression recognition},
  author={Li, Yu and Jiang, Jin and Zhu, Jianhua and Peng, Shuai and Zhou, Yuxuan and Gao, Liangcai},
  journal={Advances in Neural Information Processing Systems},
  volume={38},
  pages={129040--129074},
  year={2026}
}

@article{gshard,
  title={Gshard: Scaling giant models with conditional computation and automatic sharding},
  author={Lepikhin, Dmitry and Lee, HyoukJoong and Xu, Yuanzhong and Chen, Dehao and Firat, Orhan and Huang, Yanping and Krikun, Maxim and Shazeer, Noam and Chen, Zhifeng},
  journal={arXiv preprint arXiv:2006.16668},
  year={2020}
}

@article{switch_transformer,
  title={Switch transformers: Scaling to trillion parameter models with simple and efficient sparsity},
  author={Fedus, William and Zoph, Barret and Shazeer, Noam},
  journal={Journal of Machine Learning Research},
  volume={23},
  number={120},
  pages={1--39},
  year={2022}
}

@inproceedings{deepseekmoe,
  title={Deepseekmoe: Towards ultimate expert specialization in mixture-of-experts language models},
  author={Dai, Damai and Deng, Chengqi and Zhao, Chenggang and Xu, RX and Gao, Huazuo and Chen, Deli and Li, Jiashi and Zeng, Wangding and Yu, Xingkai and Wu, Yu and others},
  booktitle={Proceedings of the 62nd annual meeting of the association for computational linguistics (volume 1: Long papers)},
  pages={1280--1297},
  year={2024}
}

@article{sparse_upcycling,
  title={Sparse upcycling: Training mixture-of-experts from dense checkpoints},
  author={Komatsuzaki, Aran and Puigcerver, Joan and Lee-Thorp, James and Ruiz, Carlos Riquelme and Mustafa, Basil and Ainslie, Joshua and Tay, Yi and Dehghani, Mostafa and Houlsby, Neil},
  journal={arXiv preprint arXiv:2212.05055},
  year={2022}
}

@inproceedings{casia_hwdb,
  title={CASIA online and offline Chinese handwriting databases},
  author={Liu, Cheng-Lin and Yin, Fei and Wang, Da-Han and Wang, Qiu-Feng},
  booktitle={2011 international conference on document analysis and recognition},
  pages={37--41},
  year={2011},
  organization={IEEE}
}

@misc{rimes_complete,
  author       = {Grosicki, Emmanu{\`e}le and Carr{\'e}, Matthieu and Geoffrois, Edouard and Augustin, Emmanuel and Preteux, Fran{\c{c}}oise and Messina, Ronaldo},
  title        = {{RIMES}, Complete},
  year         = {2024},
  howpublished = {Zenodo},
  doi          = {10.5281/zenodo.10812725}
}

@article{handwriting_synthesis_survey,
  title={A survey of handwriting synthesis from 2019 to 2024: A comprehensive review},
  author={Diaz, Moises and Mendoza-Garc{\'\i}a, Andrea and Ferrer, Miguel A and Sabourin, Robert},
  journal={Pattern Recognition},
  volume={162},
  pages={111357},
  year={2025},
  publisher={Elsevier}
}

@article{fwgan,
  title={FW-GAN: frequency-driven handwriting synthesis with wave-modulated MLP generator},
  author={Tong, Huynh Tong Dang and Nam, Dang Hoai and Le Duy, Vo Nguyen},
  journal={Expert Systems with Applications},
  pages={130175},
  year={2025},
  publisher={Elsevier}
}

@article{writevit,
  title={WriteViT: Handwritten Text Generation with Vision Transformer},
  author={Nam, Dang Hoai and Khoa, Huynh Tong Dang and Le Duy, Vo Nguyen},
  journal={Expert Systems with Applications},
  pages={133742},
  year={2026},
  publisher={Elsevier}
}

@inproceedings{emuru,
  title={Zero-shot styled text image generation, but make it autoregressive},
  author={Pippi, Vittorio and Quattrini, Fabio and Cascianelli, Silvia and Tonioni, Alessio and Cucchiara, Rita},
  booktitle={2025 IEEE/CVF Conference on Computer Vision and Pattern Recognition (CVPR)},
  pages={7910--7919},
  year={2025},
  organization={IEEE}
}

@article{marti2002iam,
  title={The IAM-database: an English sentence database for offline handwriting recognition},
  author={Marti, U-V and Bunke, Horst},
  journal={International journal on document analysis and recognition},
  volume={5},
  number={1},
  pages={39--46},
  year={2002},
  publisher={Springer}
}

@inproceedings{quo_vadis_htg,
  title={Quo Vadis Handwritten Text Generation for Handwritten Text Recognition?},
  author={Pippi, Vittorio and Nikolaidou, Konstantina and Cascianelli, Silvia and Retsinas, George and Sfikas, Giorgos and Cucchiara, Rita and Liwicki, Marcus},
  booktitle={2025 IEEE/CVF International Conference on Computer Vision Workshops (ICCVW)},
  pages={7523--7533},
  year={2025},
  organization={IEEE}
}

@article{caruana1997multitask,
  title   = {Multitask Learning},
  author  = {Caruana, Rich},
  journal = {Machine Learning},
  volume  = {28},
  number  = {1},
  pages   = {41--75},
  year    = {1997}
}

@inproceedings{convnext,
  title     = {A {ConvNet} for the 2020s},
  author    = {Liu, Zhuang and Mao, Hanzi and Wu, Chao-Yuan and Feichtenhofer, Christoph and Darrell, Trevor and Xie, Saining},
  booktitle = {Proceedings of the IEEE/CVF Conference on Computer Vision and Pattern Recognition (CVPR)},
  pages     = {11976--11986},
  year      = {2022}
}

@article{mixtral,
  title   = {Mixtral of Experts},
  author  = {Jiang, Albert Q. and Sablayrolles, Alexandre and Roux, Antoine and Mensch, Arthur and Savary, Blanche and Bamford, Chris and others},
  journal = {arXiv preprint arXiv:2401.04088},
  year    = {2024}
}

@article{qwen2vl,
  title   = {{Qwen2-VL}: Enhancing Vision-Language Model's Perception of the World at Any Resolution},
  author  = {Wang, Peng and Bai, Shuai and Tan, Sinan and Wang, Shijie and Fan, Zhihao and Bai, Jinze and others},
  journal = {arXiv preprint arXiv:2409.12191},
  year    = {2024}
}

@inproceedings{monkey,
  title     = {Monkey: Image Resolution and Text Label Are Important Things for Large Multi-modal Models},
  author    = {Li, Zhang and Yang, Biao and Liu, Qiang and Ma, Zhiyin and Zhang, Shuo and Yang, Jingxu and Sun, Yabo and Liu, Yuliang and Bai, Xiang},
  booktitle = {Proceedings of the IEEE/CVF Conference on Computer Vision and Pattern Recognition (CVPR)},
  pages     = {26763--26773},
  year      = {2024}
}

@inproceedings{flamingo,
  title     = {Flamingo: a Visual Language Model for Few-Shot Learning},
  author    = {Alayrac, Jean-Baptiste and Donahue, Jeff and Luc, Pauline and Miech, Antoine and Barr, Iain and Hasson, Yana and others},
  booktitle = {Advances in Neural Information Processing Systems (NeurIPS)},
  volume    = {35},
  pages     = {23716--23736},
  year      = {2022}
}

@inproceedings{deformable_detr,
  title     = {Deformable {DETR}: Deformable Transformers for End-to-End Object Detection},
  author    = {Zhu, Xizhou and Su, Weijie and Lu, Lewei and Li, Bin and Wang, Xiaogang and Dai, Jifeng},
  booktitle = {International Conference on Learning Representations (ICLR)},
  year      = {2021}
}

@inproceedings{fpn,
  title     = {Feature Pyramid Networks for Object Detection},
  author    = {Lin, Tsung-Yi and Doll{\'a}r, Piotr and Girshick, Ross and He, Kaiming and Hariharan, Bharath and Belongie, Serge},
  booktitle = {Proceedings of the IEEE Conference on Computer Vision and Pattern Recognition (CVPR)},
  pages     = {2117--2125},
  year      = {2017}
}

\end{document}